\documentclass[10pt,twocolumn,letterpaper]{article}

\usepackage[pagenumbers]{cvpr} 

\definecolor{cvprblue}{rgb}{0.21,0.49,0.74}
\usepackage[pagebackref,breaklinks,colorlinks,allcolors=cvprblue]{hyperref}

\usepackage{graphicx}
\usepackage{colortbl}
\usepackage{listings}
\usepackage{booktabs}
\usepackage{multirow}
\usepackage{array}
\usepackage{colortbl}
\usepackage{makecell}
\usepackage{quoting}
\usepackage{framed}
\usepackage{arydshln}
\usepackage{longtable}
\usepackage{ltablex}

\usepackage{wrapfig}
\usepackage{booktabs}
\usepackage{graphicx}
\usepackage{pifont}
\usepackage{multirow}
\newcolumntype{M}{>{\centering\arraybackslash}m{45pt}}
            
\usepackage{url}
\usepackage[utf8]{inputenc} 
\usepackage[T1]{fontenc}    

\usepackage{url}            
\usepackage{booktabs}       
\usepackage{amsfonts}       
\usepackage{nicefrac}       
\usepackage{microtype}      

\usepackage{framed}
\usepackage{threeparttable}
\usepackage{multirow}
\usepackage{graphicx}

\usepackage{newtxmath}
\usepackage{rotating}

\usepackage{wrapfig}
\usepackage{enumitem}
\usepackage{ragged2e}
\usepackage{subcaption}

\usepackage{CJKutf8}         

\usepackage{fontawesome5} 
\usepackage{academicons}  
\usepackage{xcolor}       
\usepackage{cuted}

\usepackage{threeparttable}
\usepackage{multirow}
\usepackage{graphicx}
\usepackage{newtxmath}
\usepackage{rotating}

\usepackage{wrapfig}
\usepackage{enumitem}
\usepackage{ragged2e}
\usepackage{listings}

\usepackage{arydshln}
\usepackage{makecell}

\usepackage[capitalize,noabbrev]{cleveref}
\usepackage{lipsum}
\usepackage{booktabs}
\usepackage{multirow}
\usepackage{array}
\usepackage{arydshln}
\usepackage{colortbl}
\usepackage{fontawesome5}
\usepackage[utf8]{inputenc}
\usepackage{tabularx}   
\usepackage{amsmath}    
\usepackage{pifont}      
\usepackage{adjustbox}

\newcommand{\cmark}{\ding{51}\xspace}%
\newcommand{\xmark}{\ding{55}\xspace}%
\newcommand{\xmarkg}{\textcolor{lightgray}{\ding{55}}\xspace}

\definecolor{oursbg}{HTML}{EAF2F8}

\definecolor{tabblue}{HTML}{1f77b4}
\definecolor{taborange}{HTML}{ff7f0e}
\definecolor{tabgreen}{HTML}{2ca02c}
\definecolor{tabred}{HTML}{d62728}
\definecolor{tabpurple}{HTML}{9467bd}
\definecolor{tabpink}{HTML}{ff0080}
\newcolumntype{C}{>{\centering\arraybackslash}X}
\definecolor{darkred}{RGB}{139,0,0}
\definecolor{SATTbg}{RGB}{250,244,244}

\def\confName{CVPR}
\def\confYear{2027}

\title{Towards 360° Embodied UAV Tracking Without Turning}

\author{
\begin{tabular}{c}
Bingyu Li$^{1}$ \thanks{Work done during an internship at TeleAI.} \quad Wu Du$^{2}$  \quad Zhiyuan Zhao$^{3}$ \quad Junyu Gao$^{3, 2}$ \quad Xuelong Li$^{3}$ \thanks{Corresponding Author}\\
$^{1}$University of Science and Technology of China\quad
$^{2}$Northwestern Polytechnical University \\
$^{3}$The Institute of Artificial Intelligence (TeleAI) of China Telecom
\end{tabular}
}

\begin{document}
\maketitle

\begin{strip}
\vspace{-15mm}
\centering
\includegraphics[width=\linewidth]{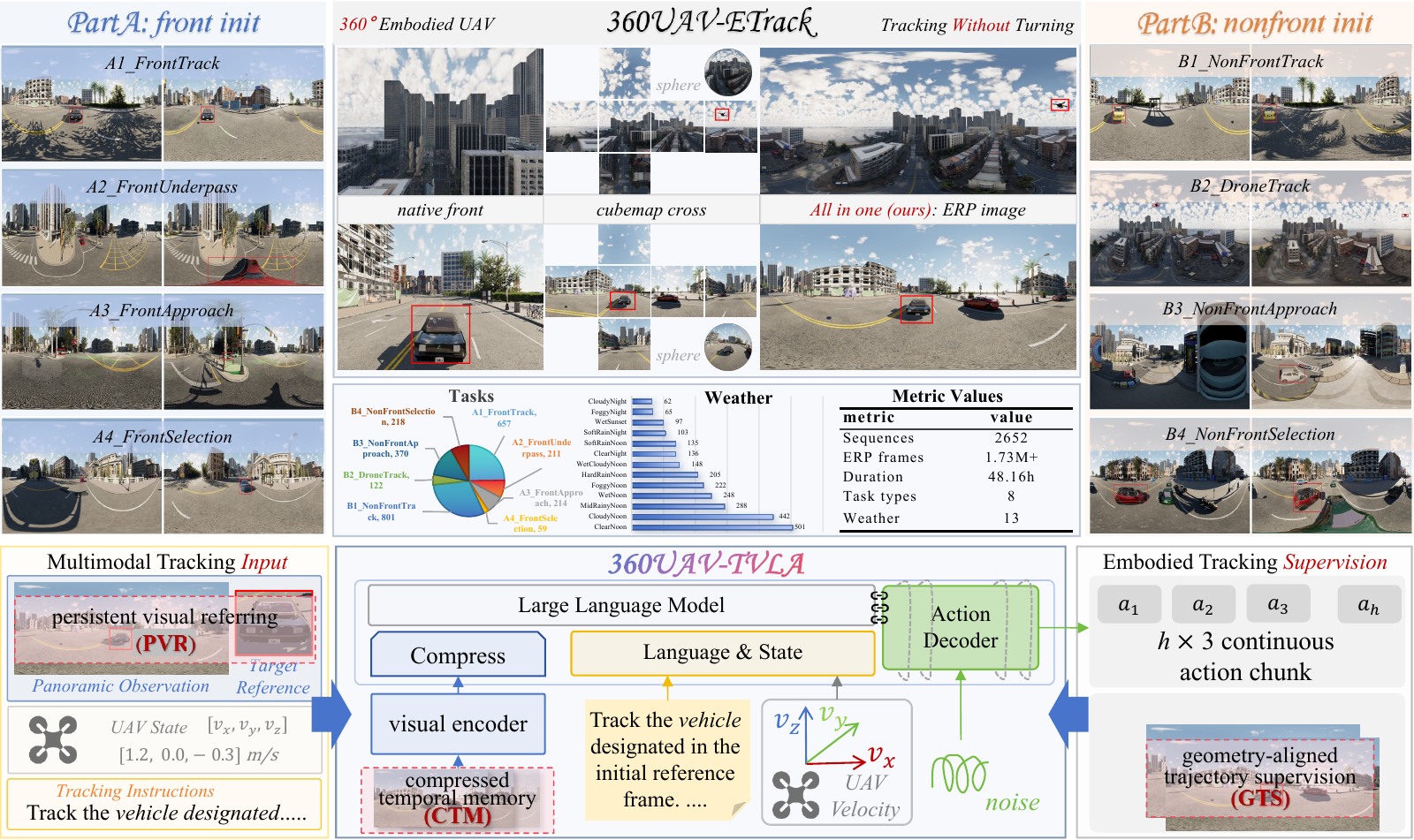}

\vspace{-2mm}

\captionof{figure}{
\textbf{Overview of 360UAV-ETrack and 360UAV-TVLA.}
The upper panel presents the benchmark scenarios, panoramic
observations, and dataset statistics, while the lower
panel illustrates our VLA framework for
full-spherical embodied UAV tracking.
}

\label{fig:paper_overall}
\end{strip}

\begin{abstract}
Conventional embodied UAV tracking couples target observability
with heading control because targets may leave the field of view
of forward-facing cameras.
We investigate a complementary paradigm that uses full-spherical
perception to follow designated targets without active heading adjustment.
To support this setting, we introduce 360UAV-ETrack,
a simulation benchmark comprising 2,652 sequences,
over 1.73 million synchronized observations, eight
tracking tasks, and 150 held-out closed-loop scenarios.
We propose 360UAV-TVLA, a panoramic vision--language--action
(VLA) framework integrating persistent visual referring (PVR),
compressed temporal memory (CTM), and geometry-aligned
trajectory supervision (GTS) for instance identification,
motion reasoning, and three-dimensional control.
Experiments across two VLA backbones show consistent gains
over standard fine-tuning, with normalized average tracking
frames increasing from 22.56\% to 35.28\% for $\pi_0$
and from 25.68\% to 33.64\% for $\pi_{0.5}$.
Our benchmark further reveals a substantial gap between
offline imitation accuracy and closed-loop tracking
reliability, establishing a challenging testbed for
future research\footnote{%
\textcolor{tabpink}{%
\url{https://libingyu01.github.io/360UAV-ETrack_360UAV-TVLA/}}}.
\end{abstract}

\section{Introduction}
\label{sec:introduction}

Embodied UAV tracking requires an agent to continuously
observe and follow a designated moving target.
Unlike visual object tracking in prerecorded videos,
embodied tracking operates in a closed perception--action
loop, where the agent's actions influence subsequent
observations and its spatial relationship with the
target~\citep{wang2025trackvla}.
Existing approaches predominantly rely on forward-facing
perspective cameras. Consequently, targets moving to the
side or behind the UAV may leave the field of view (FoV),
even when they remain nearby.
Maintaining visual contact thus requires active heading
adjustment, coupling target observability with heading.

\begin{figure}[t]
    \centering
    \includegraphics[width=\linewidth]{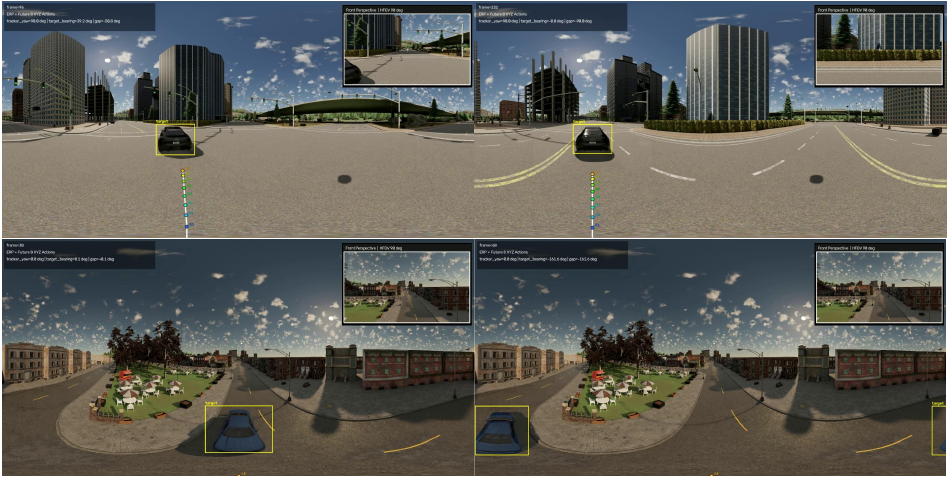}
    \vspace{-6.5mm}
    \caption{\textbf{Motivation for full-spherical embodied tracking.}
    ERP observations retain visual access to targets beyond the frontal FoV.}
    \label{fig:tracking_motivation}
    \vspace{-3.5mm}
\end{figure}

We investigate a complementary tracking paradigm that
decouples angular target coverage from heading control
through full-spherical perception.
As illustrated in Fig.~\ref{fig:tracking_motivation},
equirectangular projection (ERP) extends visual coverage
to lateral and rear directions without camera reorientation.
This motivates \emph{tracking without turning}, in which
the UAV maintains its initial heading and follows the
designated target exclusively through three-dimensional
translational control.
While physical occlusions remain unavoidable, full-spherical
perception alleviates target loss caused by limited FoV.

Despite its potential, full-spherical perception remains
underexplored in embodied UAV tracking.
Panoramic tracking benchmarks such as
360VOT~\citep{huang2023_360vot} and
360VOTS~\citep{xu2025_360vots} focus on offline target
localization, whereas embodied approaches such as
TrackVLA~\citep{wang2025trackvla} and
UAV-Track VLA~\citep{zhang2026uavtrackvla}
primarily rely on perspective observations.
As summarized in Table~\ref{tab:dataset_comparison},
existing benchmarks provide limited support for
full-spherical closed-loop UAV tracking.

To support systematic research in this setting, we
introduce \textbf{360UAV-ETrack}, a large-scale simulation
benchmark built in CARLA.
As illustrated in Fig.~\ref{fig:paper_overall},
it provides synchronized panoramic and perspective
observations, expert demonstrations, and diverse tracking
scenarios spanning frontal and non-frontal target
configurations.
The training and validation collection comprises
2,652 sequences and 1,733,766 synchronized observation
steps, complemented by 150 held-out scenarios for closed-loop evaluation.

However, full-spherical perception introduces several
challenges beyond expanded visual coverage.
The wide FoV increases instance ambiguity, making
\textit{target identity} difficult to maintain across
viewing directions.
A single panoramic observation provides limited evidence
of \textit{temporal motion}, while ERP distortions and
projection discontinuities complicate the translation of
\textit{spatial relationships} into metric control commands.
To address these challenges, we propose \textbf{360UAV-TVLA},
a reference-conditioned VLA framework comprising three
complementary components.
Specifically, persistent visual referring (PVR) establishes an
instance-specific anchor to preserve target identity across
viewing directions.
Meanwhile, compressed temporal memory (CTM) efficiently aggregates
historical panoramic features to capture target dynamics.
Furthermore, geometry-aligned trajectory supervision (GTS) predicts
future target positions in the fixed-heading control coordinate
system, explicitly bridging panoramic perception and velocity
control.

Experiments across two VLA backbones demonstrate consistent
improvements over standard fine-tuning, increasing normalized
ATF from 22.56\% to 35.28\% with $\pi_0$ and from 25.68\%
to 33.64\% with $\pi_{0.5}$.
The results also reveal a substantial discrepancy between
offline imitation accuracy and online tracking reliability,
highlighting the challenges of panoramic embodied tracking
and motivating further research in this direction.

Our contributions are summarized as follows:
\begin{itemize}
    \item We investigate fixed-heading embodied UAV tracking,
    decoupling angular target coverage from heading control
    through full-spherical perception.

    \item To support this paradigm, we introduce 360UAV-ETrack,
    a large-scale benchmark featuring diverse tracking tasks,
    expert demonstrations, and closed-loop evaluation.

    \item We develop 360UAV-TVLA, a reference-conditioned VLA
    framework with compressed panoramic memory and control-aligned
    trajectory supervision, demonstrating improved closed-loop
    tracking across two VLA backbones.
\end{itemize}


\providecommand{\PanoDatasetName}{360UAV-ETrack}
\providecommand{\cmark}{\ensuremath{\checkmark}}
\providecommand{\xmark}{\ensuremath{\times}}

\begin{table*}[t]
    \centering
    \caption{
    Comparison with representative visual tracking,
    embodied tracking, and aerial navigation benchmarks.
    Our benchmark supports full-spherical perception,
    expert action supervision, and closed-loop tracking
    under fixed-heading control.
    }
    \label{tab:dataset_comparison}
    \vspace{-1em}
    \small
    \setlength{\tabcolsep}{3.5pt}
    \renewcommand{\arraystretch}{0.8}

    \resizebox{\textwidth}{!}{%
    \begin{tabular}{@{}lllccccclll@{}}
        \toprule
        Dataset / Benchmark
        & Platform
        & Observation
        & Visual Reference
        & Language
        & Degrees of Freedom
        & Closed-Loop
        & Fixed Heading
        & Sequences
        & Data Scale
        & Data Source \\
        \midrule

        \multicolumn{11}{@{}l}{
            \textit{Visual Tracking Benchmarks}
        } \\
        \addlinespace[2pt]

        WebUAV-3M~\cite{zhang2022webuav}
        & UAV & Perspective
        & \cmark & \cmark & --
        & \xmark & --
        & 4,500 & 3,300,000 & Real-world \\

        360VOT~\cite{huang2023_360vot}
        & Various & Panorama
        & \cmark & \xmark & --
        & \xmark & --
        & 120 & 113,000 & Real-world \\

        QuadTrack~\cite{luo2025omnidirectional}
        & Quadruped & Panorama
        & -- & \xmark & --
        & \xmark & --
        & 32 & 19,200 & Real-world \\

        \midrule

        \multicolumn{11}{@{}l}{
            \textit{Embodied Tracking Benchmarks}
        } \\
        \addlinespace[2pt]

        EVT-Bench~\cite{wang2025trackvla}
        & Ground Robot & Perspective
        & \xmark & \cmark & 3
        & \cmark & --
        & 25,986 & 855,000 & Hybrid \\

        UAV-Track~\cite{zhang2026uavtrackvla}
        & UAV & Perspective
        & \xmark & \cmark & 4
        & \cmark & --
        & Not reported & 892,800 & Simulation \\

        \midrule

        \multicolumn{11}{@{}l}{
            \textit{Aerial Navigation Benchmarks}
        } \\
        \addlinespace[2pt]

        AVDN~\cite{fan2023aerial}
        & UAV & Top-down imagery
        & \xmark & \cmark & 4
        & \cmark & --
        & 3,064 & -- & Reconstructed \\

        AerialVLN~\cite{liu2023aerialvln}
        & UAV & Perspective
        & \xmark & \cmark & 4
        & \cmark & --
        & 8,446 & -- & Simulation \\

        CityNav~\cite{lee2025citynav}
        & UAV & Perspective
        & \xmark & \cmark & 4
        & \cmark & --
        & 32,637 & -- & Reconstructed \\

        OpenUAV~\cite{wang2025towards}
        & UAV & Multiple perspective views
        & \xmark & \cmark & 6
        & \cmark & --
        & 12,149 & -- & Simulation \\

        OpenFly~\cite{gao2026openfly}
        & UAV & Perspective
        & \xmark & \cmark & 4
        & \cmark & --
        & 100,000 & -- & Hybrid \\

        \midrule

        \multicolumn{11}{@{}l}{
            \textit{Full-Spherical Embodied UAV Tracking}
        } \\
        \addlinespace[2pt]

        \textbf{\PanoDatasetName}
        & UAV & Panorama + perspective
        & \cmark & \cmark & \textbf{3}
        & \cmark & \cmark
        & 2,652 & 1,733,766 & Simulation \\

        \bottomrule
    \end{tabular}%
    }
    \vspace{-2.5mm}
\end{table*}

\section{Related Work}

\subsection{Visual Perception and Tracking}
Panoramic vision expands scene coverage but introduces
projection distortions that motivate geometry-aware
representations~\citep{cohen2018spherical}.
360VOT~\citep{huang2023_360vot} and
360VOTS~\citep{xu2025_360vots} benchmark omnidirectional
object tracking and segmentation, while
PanoVOS~\citep{yan2024panovos} investigates temporal
consistency in panoramic video segmentation.
Recent efforts extend panoramic perception to
multi-object tracking~\citep{luo2025omnidirectional,deng2026cylindtrack},
out-of-view point tracking~\citep{hudson2026tapvid},
and spatial reasoning~\citep{lin2026panoenv}.
However, these studies primarily address perception
rather than action-conditioned target following,
leaving full-spherical closed-loop tracking
insufficiently explored.

\subsection{Embodied Visual Tracking}
Embodied visual tracking couples target perception
with active motion control.
Early methods improve robustness to distractors
through adversarial learning~\citep{zhong2021advatplus,zhong2021distraction,AIFlowAn,AIFlowShao},
while offline reinforcement learning and adaptive
policies support robust tracking across
settings~\citep{zhong2024offlineevt,wu2026adatracker}.
UAV-oriented methods such as D-VAT~\citep{dionigi2024dvat},
DAT~\citep{sun2025openworld}, and
OA-VAT~\citep{sun2026oavat} investigate autonomous
following and occlusion recovery.
Recent VLA approaches, including
TrackVLA~\citep{wang2025trackvla},
UAV-Track VLA~\citep{zhang2026uavtrackvla}, and
TrackVLA++~\citep{liu2026trackvlapp},
unify target understanding and action prediction;
VLM-assisted tracking further explores failure
recovery~\citep{wu2025vlmassistant}.
CosFly-Track~\citep{wang2026cosflytrack} provides
large-scale demonstrations for aerial following.
Nevertheless, these approaches predominantly use
perspective observations, leaving the role of
full-spherical perception in fixed-heading
closed-loop UAV tracking underexplored.

\section{360UAV-ETrack Benchmark}
\label{sec:benchmark}

\subsection{Simulation and Data Collection}
\label{sec:simulation}

We construct \textbf{360UAV-ETrack} in CARLA to study
full-spherical embodied UAV tracking under fixed-heading
control. As illustrated in Fig.~\ref{fig:paper_overall},
the UAV follows a designated moving target through
three-dimensional translation without heading adjustments.
The training and validation set contains 2,652 sequences
with 1,733,766 ERP frames (48.16 hours), covering eight
tracking tasks and 13 weather conditions.
Demonstrations are collected across five towns:
Town02, Town05, Town06, Town07, and Town10HD.

The UAV captures $1920\times960$ equirectangular
projection (ERP) observations at $10\,\mathrm{Hz}$,
while simulation and low-level control operate at
$30\,\mathrm{Hz}$.
Each demonstration provides synchronized panoramic
and perspective observations, target references and
annotations, UAV motion states, and expert velocity
commands.

\subsection{Tracking Tasks}
\label{sec:tracking_tasks}

As shown in Fig.~\ref{fig:paper_overall},
360UAV-ETrack comprises eight tasks organized into
two evaluation settings.

\textbf{Task I: Front-Initialized Tracking.}
Targets are initialized within the frontal
$90^\circ$ field of view.
The four tasks are (A1) FrontTrack,
(A2) FrontUnderpass, (A3) FrontApproach,
and (A4) FrontSelection.
They evaluate persistent following, underpass
navigation, target approach, and identity preservation
among distractors, respectively.

\textbf{Task II: Omnidirectional Acquisition and Tracking.}
The four tasks are (B1) NonFrontTrack,
(B2) DroneTrack, (B3) NonFrontApproach,
and (B4) NonFrontSelection.
These tasks extend evaluation to non-frontal target
configurations and aerial targets, testing target
acquisition and subsequent tracking under
full-spherical observations.

\textbf{Held-Out Evaluation Set.}
We construct 150 frozen evaluation episodes across
Town01, Town03, and Town04, which are disjoint from
the training towns.
Task I contains 51 episodes, with 21/7/5/18 episodes
for A1--A4.
Task II contains 99 episodes, with 44/6/13/36 episodes
for B1--B4.
The set includes 72 pedestrian, 72 vehicle, and
6 UAV-target episodes.
Recorded expert demonstrations support offline replay,
while the corresponding frozen scenarios support
online closed-loop evaluation.

\begin{figure*}[t]
    \centering
    \includegraphics[width=0.96\textwidth]{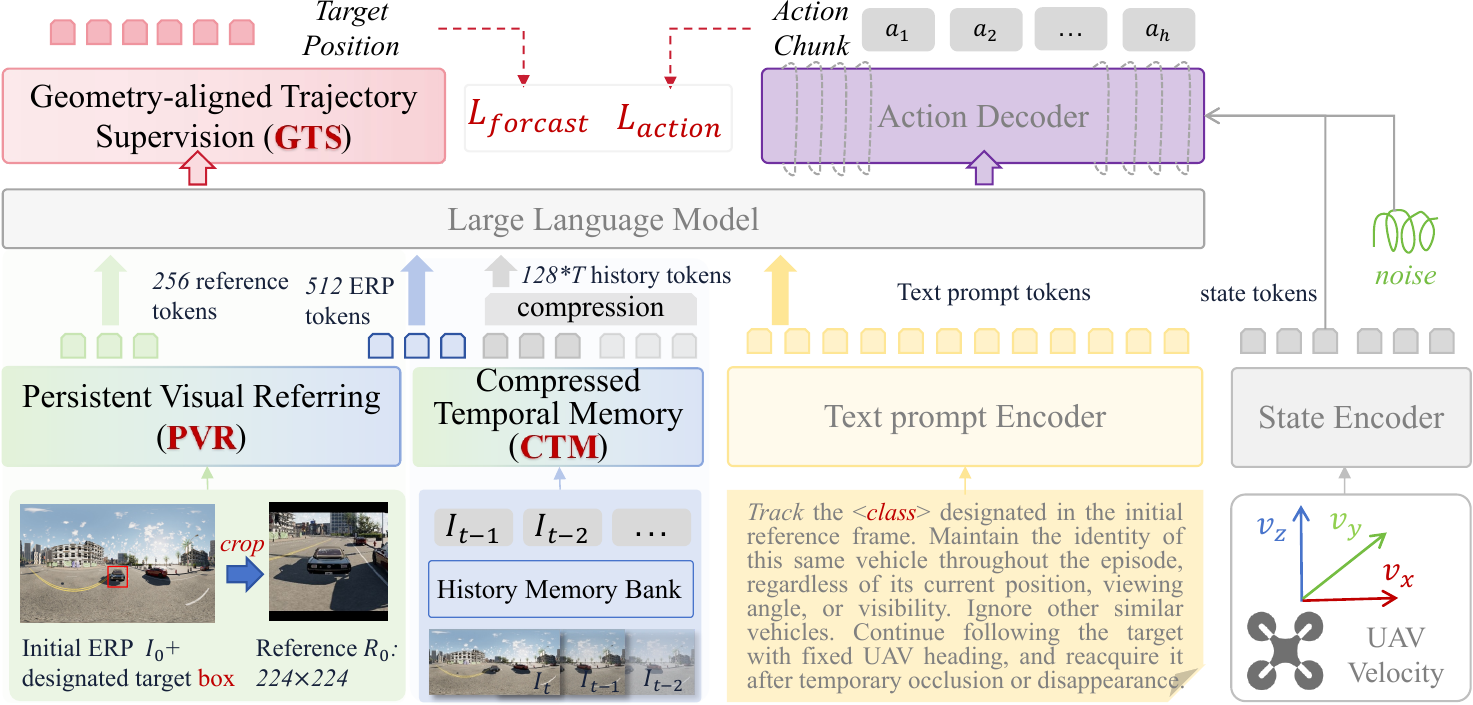}
    \vspace{-0.6em}
    \caption{
    \textbf{Overview of 360UAV-TVLA.} A training-only auxiliary head
    predicts 10 future target positions relative to the current UAV.
    At inference, only the first predicted action is executed
    before replanning under fixed-heading control.
    }
    \label{fig:Fig_3_method}
    \vspace{-1.0em}
\end{figure*}

\subsection{Evaluation Protocol}
\label{sec:evaluation}

All policies predict three-dimensional velocity
commands in the same heading-aligned coordinate frame,
using the first predicted command at each decision step.
We evaluate offline prediction and online tracking
separately.

\textbf{Offline Expert Replay.}
At each timestamp, the policy receives the recorded
expert observation and motion-state inputs.
Predicted velocities are integrated from the expert
initial position with $\Delta t=0.1\,\mathrm{s}$,
without feeding the integrated trajectory back into
the recorded inputs.
We report average displacement error (ADE) against
the expert trajectory, averaging position errors over
each complete sequence and then across episodes.
Offline success rate, $\mathrm{SR}_{\mathrm{off}}$,
evaluates the integrated trajectory against the
recorded target positions using the original distance
constraints and a 15-consecutive-frame failure threshold.
It measures distance compliance and does not evaluate
visibility from the predicted positions.

\textbf{Online Closed-Loop Evaluation.}
Each policy receives simulator observations and executes
velocity commands in closed loop. We compare three timing
protocols by the delay included in simulated time:
\begin{equation}
D_{\mathrm{(i)}} = 0,
D_{\mathrm{(ii)}} = L_{\mathrm{model}} + L_{\mathrm{ext}},
D_{\mathrm{(iii)}} = L_{\mathrm{model}},
\end{equation}
where $L_{\mathrm{model}}$ is the measured policy latency,
from policy entry to the complete action chunk being
available as a synchronized CPU array, and $L_{\mathrm{ext}}$
denotes external overhead, including observation preparation,
communication, and rendering.
These delays exclude physics-step quantization.

\textit{(i) Synchronous closed loop.}
The simulator pauses during inference and then advances
by $0.1\,\mathrm{s}$ (10Hz) under the new command, excluding
inference latency from simulated time.

\textit{(ii) End-to-end real-time closed loop.}
The simulator advances continuously during asynchronous
inference, exposing the policy to both model latency
and external overhead.

\textit{(iii) Model-compute-delay closed loop.}
Only measured model latency is injected into simulated time.
For an observation and UAV state captured at time $t$,
the new command takes effect at the first $30\,\mathrm{Hz}$
physics boundary at or after $t+L_{\mathrm{model}}$.
Observations and scoring run at $10\,\mathrm{Hz}$, with
at most one prediction pending.

All reported online results use protocol (iii),
which incorporates measured model computation latency
while excluding simulator rendering and communication
overhead. Detailed definitions of the three timing
protocols are provided in the appendix.
\section{Method}
\label{sec:method}

\subsection{Overall Framework}
\label{sec:fs_overview}
Full-spherical perception decouples target observability
from UAV heading control, yet poses three fundamental challenges:
(1) how to identify the designated instance within wide-FoV \textit{ERP observations},
(2) how to reason about target dynamics from temporal observations for \textit{persistent tracking}, and
(3) how to \textit{effectively translate} spherical perception into three-dimensional control.
To address these challenges, we propose \textbf{360UAV-TVLA},
which integrates three complementary components:
persistent visual referring (PVR, $\mathcal R$),
compressed temporal memory (CTM, $\mathcal M$),
and geometry-aligned trajectory supervision (GTS, $\mathcal G$). As illustrated in
Fig.~\ref{fig:Fig_3_method}, these components are integrated
with a $\pi$-series VLA backbone $\mathcal B_\Theta$.
At observation index $t$, the overall framework is formulated as

\begin{equation}
    \begin{aligned}
    &R=\mathcal R(I_0,\mathbf b_0^{\mathrm{init}}),\quad
    \overline{\mathcal H}_t=\mathcal M(\mathcal H_t),\\
    &(Z_t,\widehat{\mathbf A}_t)=\mathcal B_\Theta\!\left(
    I_t,R,\overline{\mathcal H}_t,\ell,\mathbf v_t;
    \boldsymbol\epsilon\right),\\
    &\widehat U_t=\mathcal G(Z_t).
    \end{aligned}
    \label{eq:fs_policy}
\end{equation}
where $I_0$ and $\mathbf b_0^{\mathrm{init}}$ denote the
initial panorama and the bounding box of the designated
target in the initial observation, respectively.
$I_t$, $\mathcal H_t$, $\ell$, and $\mathbf v_t$
denote the current panorama, historical observations,
language instruction, and measured UAV velocity, respectively.
$\boldsymbol\epsilon \sim \mathcal N(\mathbf 0,\mathbf I)$
denotes the Gaussian noise used to initialize the
action generation process.
$Z_t$ is the shared multimodal representation,
$\widehat{\mathbf A}_t\in\mathbb R^{h\times3}$ denotes
the predicted velocity chunk, and
$\widehat U_t\in\mathbb R^{h\times3}$ represents
the predicted target trajectory in normalized
control-aligned coordinates.
The prediction horizon is $h=10$, with actions sampled at
$10\,\mathrm{Hz}$ in the initial-heading-aligned frame.
The three components respectively address instance referring,
temporal reasoning, and control-aligned motion learning.
$\mathcal G$ is used exclusively during training to regularize
the shared representation without additional inference cost.

\begin{table*}[t]
\centering
\caption{
\textbf{Online closed-loop tracking performance on 360UAV-ETrack.}
Task I evaluates front-initialized tracking, while Task II
evaluates omnidirectional acquisition and tracking.
ATF is reported in frames, normalized ATF and SR in \%,
and FPS in decisions/s.
Normalized ATF averages each episode's ATF divided by its
prescribed horizon.
}
\vspace{-1em}
\label{tab:overall_embodied_tracking}
\small
\setlength{\tabcolsep}{4pt}
\renewcommand{\arraystretch}{1.00}

\resizebox{\linewidth}{!}{%
\begin{tabular}{@{}lcc*{9}{c}c@{}}
\toprule
\multirow{2}{*}{Method}
& \multirow{2}{*}{Publication}
& \multirow{2}{*}{Type}
& \multicolumn{3}{c}{
    \shortstack[c]{Task I Front-Initialized}
}
& \multicolumn{3}{c}{
    \shortstack[c]{Task II Omnidirectional}
}
& \multicolumn{3}{c}{
    \shortstack[c]{Overall Both Tasks}
}
& \multirow{2}{*}{
    \shortstack[c]{FPS$\uparrow$\\(decisions/s)}
} \\
\cmidrule(lr){4-6}
\cmidrule(lr){7-9}
\cmidrule(lr){10-12}
& & &
ATF$\uparrow$
& $\mathrm{ATF}_{\mathrm{norm}}\uparrow$
& $\mathrm{SR}_{\mathrm{on}}\uparrow$
& ATF$\uparrow$
& $\mathrm{ATF}_{\mathrm{norm}}\uparrow$
& $\mathrm{SR}_{\mathrm{on}}\uparrow$
& ATF$\uparrow$
& $\mathrm{ATF}_{\mathrm{norm}}\uparrow$
& $\mathrm{SR}_{\mathrm{on}}\uparrow$
& \\
\midrule

\multicolumn{13}{
@{}>{\columncolor{gray!10}[0pt][0pt]}l@{}
}{
\textit{Direct Inference (Without Benchmark Fine-Tuning)}
} \\

Uni-NaVid~\citep{zhang2025uninavid}
& RSS'25 & VLA
& 37.31 & 3.09 & 0.00
& 39.00 & 4.89 & 0.00
& 38.43 & 4.28 & 0.00
& 3.08 \\

OmTrackVLA~\citep{omtrackvla2025}
& GitHub'25 & VLA
& 39.31 & 3.30 & 0.00
& 40.90 & 5.12 & 0.00
& 40.36 & 4.50 & 0.00
& 1.31 \\

$\pi_0$ (w/o ft)~\citep{black2025pi0}
& RSS'25 & VLA
& 45.69 & 3.79 & 0.00
& 36.57 & 4.57 & 0.00
& 39.67 & 4.30 & 0.00
& 21.36 \\

$\pi_{0.5}$ (w/o ft)~\citep{black2025pi05}
& CoRL'25 & VLA
& 76.53 & 6.46 & 0.00
& 34.19 & 4.18 & 0.00
& 48.59 & 4.95 & 0.00
& 20.59 \\

\midrule

\multicolumn{13}{
@{}>{\columncolor{gray!10}[0pt][0pt]}l@{}
}{
\textit{Fine-Tuned Models}
} \\

Temporal BC
& Baseline & BC
& 149.10 & 12.62 & 0.00
& 98.67 & 11.80 & 2.02
& 115.81 & 12.08 & 1.33
& 372.48 \\

D-VAT-BC~\citep{dionigi2024dvat}
& RA-L'24 & VAT
& 171.33 & 14.00 & 1.96
& 85.66 & 10.90 & 1.01
& 114.79 & 11.95 & 1.33
& 232.13 \\

Fast-WAM~\citep{yuan2026fastwam}
& NeurIPS'26 & WAM
& 170.16 & 14.54 & 1.96
& 178.66 & 20.69 & 3.03
& 175.77 & 18.60 & 2.67
& 3.16 \\

$\pi_0$ (w/ ft)~\citep{black2025pi0}
& RSS'25 & VLA
& 310.73 & 23.90 & 3.92
& 188.30 & 21.87 & 6.06
& 229.93 & 22.56 & 5.33
& 20.65 \\

$\pi_{0.5}$ (w/ ft)~\citep{black2025pi05}
& CoRL'25 & VLA
& 305.55 & 24.40 & 5.88
& 212.61 & 26.34 & 7.07
& 244.21 & 25.68 & \underline{6.67}
& 19.56 \\

\midrule

\rowcolor{blue!4}
\multicolumn{1}{
@{}>{\columncolor{blue!4}[0pt][\tabcolsep]}l
}{
\textbf{360UAV-TVLA ($\pi_0$)}
}
& -- & \textbf{VLA}
& \textbf{537.84}
& \textbf{39.21}
& \underline{7.84}
& \underline{291.56}
& \underline{33.25}
& \textbf{12.12}
& \textbf{375.29}
& \textbf{35.28}
& \textbf{10.67}
& \multicolumn{1}{
>{\columncolor{blue!4}[\tabcolsep][0pt]}c@{}
}{19.56} \\

\rowcolor{blue!4}
\multicolumn{1}{
@{}>{\columncolor{blue!4}[0pt][\tabcolsep]}l
}{
\textbf{360UAV-TVLA ($\pi_{0.5}$)}
}
& -- & \textbf{VLA}
& \underline{393.14}
& \underline{30.17}
& \textbf{9.80}
& \textbf{308.39}
& \textbf{35.42}
& \underline{11.11}
& \underline{337.21}
& \underline{33.64}
& \textbf{10.67}
& \multicolumn{1}{
>{\columncolor{blue!4}[\tabcolsep][0pt]}c@{}
}{18.90} \\

\bottomrule
\end{tabular}%
}
\vspace{-1.2em}
\end{table*}
\begin{table}[t]
\centering
\caption{\textbf{Offline expert-replay performance.}
All methods receive fixed expert observations.
ADE is in (m) and SR in (\%).}
\label{tab:offline_expert_replay}
\vspace{-1.0em}
\small
\setlength{\tabcolsep}{5pt}
\renewcommand{\arraystretch}{0.95}

\resizebox{\linewidth}{!}{%
\begin{tabular}{@{}lllllll@{}}
\toprule
& \multicolumn{2}{c}{Task I (51)}
& \multicolumn{2}{c}{Task II (99)}
& \multicolumn{2}{c}{Overall (150)} \\
\cmidrule(lr){2-3}
\cmidrule(lr){4-5}
\cmidrule(l){6-7}
Method
& ADE$\downarrow$ & $\mathrm{SR}_{\mathrm{off}}\uparrow$
& ADE$\downarrow$ & $\mathrm{SR}_{\mathrm{off}}\uparrow$
& ADE$\downarrow$ & $\mathrm{SR}_{\mathrm{off}}\uparrow$ \\
\midrule

$\pi_0$ (w/o ft)
& 144.04 & 0.00
& 125.80 & 0.00
& 132.00 & 0.00 \\

$\pi_0$ (w/ ft)
& \textbf{4.21} & \textbf{76.47}
& \underline{3.15} & \textbf{92.93}
& \textbf{3.51} & \textbf{87.33} \\

\rowcolor{blue!4}
\textbf{Ours} ($\pi_0$)
& \underline{5.17} & \underline{68.63}
& \textbf{3.02} & \underline{91.92}
& \underline{3.75} & \underline{84.00} \\

\midrule

$\pi_{0.5}$ (w/o ft)
& 64.17 & \underline{0.00}
& 68.02 & 0.00
& 66.71 & 0.00 \\

$\pi_{0.5}$ (w/ ft)
& \textbf{14.61} & \textbf{43.14}
& \textbf{11.57} & \textbf{59.60}
& \textbf{12.61} & \textbf{54.00} \\

\rowcolor{blue!4}
\textbf{Ours} ($\pi_{0.5}$)
& \underline{15.55} & \textbf{43.14}
& \underline{12.81} & \underline{54.55}
& \underline{13.74} & \underline{50.67} \\

\bottomrule
\end{tabular}%
}
\vspace{-1.2em}
\end{table}

\begin{table}[ht]
\centering
\small
\caption{\textbf{Progressive component ablation.}
All variants use the same $\pi_{0.5}$ backbone and a
15-frame failure threshold.}
\vspace{-1.2em}
\label{tab:component_ablation}
\setlength{\tabcolsep}{5pt}
\renewcommand{\arraystretch}{0.95}
\resizebox{0.8\linewidth}{!}{%
\begin{tabular}{c c c c c c}
\toprule
\multirow{2}{*}{\shortstack[c]{PVR}}
& \multirow{2}{*}{\shortstack[c]{CTM}}
& \multirow{2}{*}{\shortstack[c]{GTS}}
& \multicolumn{3}{c}{Online} \\
\cmidrule(lr){4-6}
& & &
ATF$\uparrow$
& $\mathrm{ATF}_{\mathrm{norm}}\uparrow$
& $\mathrm{SR}_{\mathrm{on}}\uparrow$ \\
\midrule

\xmarkg & \xmarkg & \xmarkg
& 183.52 & 17.96 & 2.00 \\

\cmark & \xmarkg & \xmarkg
& 244.21 & 25.68 & 6.67 \\

\cmark & \cmark & \xmarkg
& 265.73 & 28.16 & \underline{10.67} \\

\cmark & \xmarkg & \cmark
& \underline{319.79}
& \underline{31.61}
& \textbf{12.00} \\

\rowcolor{blue!4}
\cmark & \cmark & \cmark
& \textbf{337.21}
& \textbf{33.64}
& \underline{10.67} \\

\bottomrule
\end{tabular}%
}
\vspace{-1.4em}
\end{table}

\subsection{Persistent Visual Referring (PVR)}
\label{sec:fs_reference}

Full-spherical perception expands target observability but
also increases instance ambiguity, as visually similar objects
may appear across different viewing directions.
Language instructions typically describe target categories
rather than uniquely distinguish individual instances.
To maintain target identity throughout closed-loop tracking,
we introduce persistent visual referring $\mathcal R$,
which constructs an instance-specific visual reference
from the initial observation:
\begin{equation}
    R=\mathcal R(I_0,\mathbf b_0^{\mathrm{init}})
    =\mathcal C_{\mathrm{wrap}}
    (I_0,\mathbf b_0^{\mathrm{init}};\gamma),
    \label{eq:fs_reference}
\end{equation}
where $\mathcal C_{\mathrm{wrap}}$ extracts a context-expanded
target region with periodic boundary handling and aspect-preserving
padding to $224\times224$, using $\gamma=2.5$.
The periodic cropping preserves target appearance across the
panoramic seam.

The reference remains fixed throughout the episode,
providing a consistent identity anchor despite changes in
target location.
Together with the language instruction, PVR assists the policy
in distinguishing the designated target from similar distractors.
Only the initial bounding box is provided; subsequent target
coordinates are unavailable to the policy.

\subsection{Compressed Temporal Memory (CTM)}
\label{sec:fs_history}

While full-spherical perception provides comprehensive spatial
coverage, a single observation offers limited evidence of target
dynamics. Persistent tracking requires reasoning about changes
in relative target position and motion direction.
However, directly incorporating historical panoramic observations
substantially increases the visual token budget.
We therefore introduce compressed temporal memory
$\mathcal M$ to capture instance-specific motion efficiently.

Let $\mathcal H_t=(I_{t-T},\ldots,I_{t-1})$ denote the
historical observation window.
CTM adopts an asymmetric temporal representation that
compresses historical visual tokens while preserving the
full spatial details of the current observation:
\begin{equation}
    \overline{\mathcal H}_t
    =\mathcal M(\mathcal H_t)
    =\operatorname{Linear}\bigl(\mathcal E_\phi(\mathcal H_t)\bigr),
    \label{eq:fs_compression}
\end{equation}
where $\mathcal E_\phi$ is the shared SigLIP encoder.
Each historical frame is independently encoded and compressed
from 512 to 128 tokens along the token dimension,
without changing the feature dimension.
The shared linear layer is initialized with $2\times2$
spatial averaging and zero bias, reducing historical
token consumption by $75\%$.
The compressed features are concatenated in order to form the temporal memory.

To associate temporal motion with the designated instance,
the fusion module $\mathcal F_\theta$ integrates
compressed memory, current panorama, and persistent target reference:
\begin{equation}
    Z_t=\mathcal F_\theta\!\left(
    [\overline{\mathcal H}_t;
    \mathcal E_\phi(I_t);
    \mathcal E_\phi(R);
    \mathcal T(\ell,\mathbf v_t)]
    \right).
    \label{eq:fs_fusion}
\end{equation}

Temporal and source embeddings are added to visual tokens
before fusion to encode observation chronology and distinguish
input sources.
$\mathcal T$ encodes the language instruction and UAV velocity.
By combining historical motion cues with detailed current-frame
observations and persistent identity information, the resulting
representation $Z_t$ supports target-specific temporal reasoning
using only $768+128T$ visual tokens.
Memory is updated at each observation step independently
of policy inference, with unavailable history masked during attention.

\subsection{Geometry-Aligned Trajectory Supervision (GTS)}
\label{sec:fs_target_future}

\paragraph{Control-aligned motion representation.}
Full-spherical observations provide omnidirectional visual
coverage, but their projected image coordinates do not
directly describe the metric spatial relationships required
for three-dimensional velocity control.
Moreover, action imitation provides only indirect supervision
of target motion through expert UAV responses.
To bridge this perception-to-control gap, GTS represents
future target trajectories in the same fixed-heading coordinate
frame as UAV velocity commands:
\begin{equation}
    U_t^\star=\operatorname{Norm}\!\left(
    [\mathbf p^T_{t+1};\ldots;\mathbf p^T_{t+h}]
    -\mathbf p^U_t\right),
    \label{eq:fs_target_labels}
\end{equation}
where $\mathbf p^T$ and $\mathbf p^U$ denote target and UAV
positions in the initial-heading-aligned frame, respectively.
The current UAV position is subtracted from each of the
$h$ future target positions to establish a common spatial origin.
$\operatorname{Norm}$ denotes per-axis normalization using
training-set statistics.

Unlike ERP-space forecasting, this representation avoids
panoramic seam discontinuities in the supervision space.
The common UAV origin further separates target displacement
from future UAV motion, allowing target localization and
motion to be represented consistently.

\paragraph{Trajectory supervision.}
To encourage the policy to learn these geometric relationships,
GTS extracts target motion features from the shared
multimodal representation through learnable query-based
attention and a two-layer GELU MLP:
\begin{equation}
    \widehat U_t=\mathcal G(Z_t)\in\mathbb R^{h\times3},
    \label{eq:fs_target_prediction}
\end{equation}
where $\widehat U_t$ denotes the predicted target trajectory
in normalized control-aligned coordinates.
We optimize the trajectory prediction using
\begin{equation}
    \mathcal L_{\mathrm{traj}}
    =\operatorname{MaskedSmoothL1}(\widehat U_t,U_t^\star),
    \label{eq:fs_target_loss}
\end{equation}
averaged over valid future positions and all three coordinate axes.
By explicitly supervising future target motion, GTS encourages
the shared representation to capture instance-specific spatial
relationships and motion dynamics, including temporarily
occluded targets when valid annotations are available.
The trajectory prediction head is used exclusively during
training and introduces no additional inference cost.

\subsection{Joint Learning and Closed-Loop Control}
\label{sec:fs_joint}

Conditioned on the shared representation $Z_t$, the action
expert of $\mathcal B_\Theta$ generates velocity chunks through
differentiable reverse-time Euler sampling from Gaussian noise.
The predicted actions are mapped to physical velocities using
training-set normalization statistics and directly supervised
by a masked $L_1$ imitation loss $\mathcal L_{\mathrm{act}}$.
We jointly optimize action imitation and geometry-aligned
target trajectory supervision:
\begin{equation}
    \mathcal L=\mathcal L_{\mathrm{act}}
    +\lambda_{\mathrm{traj}}\mathcal L_{\mathrm{traj}},
    \label{eq:fs_joint_loss}
\end{equation}
where $\lambda_{\mathrm{traj}}$ balances velocity imitation
and target-motion learning.
Through joint optimization, PVR provides consistent instance
conditioning, CTM captures temporal motion evidence, and GTS
encourages control-aligned motion representation learning.
These complementary components connect target identity,
motion understanding, and continuous velocity control
within a unified VLA framework.

At inference, the GTS head is bypassed, and the policy directly
generates velocity commands from the shared representation.
The controller executes the first predicted velocity before
replanning with updated panoramic observations.
During inference latency, the UAV retains its previous velocity
command while maintaining a fixed heading throughout tracking.
\section{Experiments}
\label{sec:experiments}

\subsection{Implementation Details}
\label{sec:exp_setup}

We evaluate Uni-NaVid~\citep{zhang2025uninavid},
OmTrackVLA~\citep{omtrackvla2025}, $\pi_0$~\citep{black2025pi0},
and $\pi_{0.5}$~\citep{black2025pi05} without benchmark adaptation,
alongside benchmark-trained Temporal BC, D-VAT-BC~\citep{dionigi2024dvat},
Fast-WAM~\citep{yuan2026fastwam}, both $\pi$ baselines, and our method.
VLA/WAM training uses scene-disjoint splits, one epoch, eight H100 GPUs,
AdamW, batch size 64, and peak learning rate $2.5\times10^{-5}$.
Tables~\ref{tab:component_ablation} and~\ref{tab:online_ablation}
use $\pi_{0.5}$.
Both 360UAV-TVLA variants use CTM with two preceding frames
and 128 tokens per past frame, and GTS with $\lambda_{\mathrm{traj}}=2$.
Checkpoints are selected by validation velocity MAE.
The GTS head is bypassed at deployment.
Normalized ATF averages each episode's valid tracking frames
divided by its prescribed horizon.
Detailed configurations and native-interface adaptations
are provided in the appendix.


\begin{table}[t]
\centering
\small
\caption{\textbf{Online ablation results.}
The left block compares history lengths, while the right block
varies the target-auxiliary loss weight.}
\label{tab:online_ablation}

\vspace{-1em}
\setlength{\tabcolsep}{3pt}
\renewcommand{\arraystretch}{0.95}

\resizebox{\linewidth}{!}{%
\begin{tabular}{
@{}cccccc
@{\hspace{12pt}\vrule\hspace{12pt}}
cccc@{}
}
\toprule
\multicolumn{6}{c}{(a) CTM history length}
& \multicolumn{4}{c}{(b) GTS loss weight} \\
\cmidrule(lr){1-6}
\cmidrule(lr){7-10}

$T$
& $N_{\mathrm{vis}}$
& ATF$\uparrow$
& $\mathrm{ATF}_{\mathrm{norm}}\uparrow$
& SR$\uparrow$
& FPS$\uparrow$
& $\lambda_{\mathrm{traj}}$
& ATF$\uparrow$
& $\mathrm{ATF}_{\mathrm{norm}}\uparrow$
& SR$\uparrow$ \\
\midrule

0
& 768
& 319.79
& 31.61
& \textbf{12.00}
& \textbf{20.71}
& 0.01
& 261.44
& 26.96
& 6.67 \\

1
& 896
& 243.97
& 24.80
& 6.67
& \underline{19.17}
& 0.1
& 321.47
& 32.53
& \underline{8.67} \\

\multicolumn{1}{
@{}>{\columncolor{blue!4}[0pt][\tabcolsep]}c
}{2}
& \cellcolor{blue!4}1024
& \cellcolor{blue!4}\textbf{337.21}
& \cellcolor{blue!4}\textbf{33.64}
& \cellcolor{blue!4}\underline{10.67}
& \multicolumn{1}{
>{\columncolor{blue!4}[\tabcolsep][0pt]}c
@{\hspace{12pt}\vrule\hspace{12pt}}
}{18.90}
& 0.5
& 270.44
& 27.87
& 7.33 \\

4
& 1280
& 280.04
& 28.15
& 6.67
& 17.11
& \multicolumn{1}{
@{}>{\columncolor{blue!4}[0pt][\tabcolsep]}c
}{2}
& \cellcolor{blue!4}\textbf{337.21}
& \cellcolor{blue!4}\textbf{33.64}
& \multicolumn{1}{
>{\columncolor{blue!4}[\tabcolsep][0pt]}c@{}
}{\textbf{10.67}} \\

8
& 1792
& \underline{324.69}
& \underline{32.91}
& 8.67
& 13.79
& 5
& \underline{325.75}
& \underline{32.65}
& 6.67 \\

\bottomrule
\end{tabular}%
}

\vspace{-1.0em}
\end{table}
\begin{table}[htbp]
\centering
\small
\caption{\textbf{Full-spherical observation vs. active steering.}
We compare (A) perspective observation with fixed-heading
XYZ control, (B) ERP observation with fixed-heading XYZ
control, and (C) perspective observation with active
XYZ--yaw control.}
\vspace{-1.2em}
\label{tab:erp_fixed_heading_ablation}
\setlength{\tabcolsep}{2pt}
\renewcommand{\arraystretch}{0.95}
\resizebox{\linewidth}{!}{%
\begin{tabular}{l c c cccc}
\toprule
\multirow{2}{*}{Method}
& \multirow{2}{*}{Observation}
& \multirow{2}{*}{Action}
& \multicolumn{2}{c}{Task I}
& \multicolumn{2}{c}{Task II} \\
\cmidrule(lr){4-5}
\cmidrule(lr){6-7}
& & &
ATF$\uparrow$
& $\mathrm{SR}_{\mathrm{on}}\uparrow$
& ATF$\uparrow$
& $\mathrm{SR}_{\mathrm{on}}\uparrow$ \\
\midrule

A: Limited FoV ($\pi_{0.5}$)
& PER $90^\circ$
& $(v_x,v_y,v_z)$
& 265.08 & 3.92
& 181.54 & 5.05 \\

\rowcolor{blue!4}
\textbf{B: Ours} ($\pi_{0.5}$)
& ERP $360^\circ$
& $(v_x,v_y,v_z)$
& \underline{393.14} & \textbf{9.80}
& \textbf{308.39} & \underline{11.11} \\

\rowcolor{blue!4}
\textbf{B: Ours} ($\pi_{0}$)
& ERP $360^\circ$
& $(v_x,v_y,v_z)$
& \textbf{537.84} & \underline{7.84}
& \underline{291.56} & \textbf{12.12} \\

C: Active steering ($\pi_{0.5}$)
& PER $90^\circ$
& $(v_x,v_y,v_z,\omega_{\mathrm{yaw}})$
& 164.35 & 0.00
& 69.35 & 1.01 \\

\bottomrule
\end{tabular}%
}
\vspace{-0.8em}
\end{table}

\subsection{Overall Benchmark Performance}
\label{sec:exp_overall}

Table~\ref{tab:overall_embodied_tracking} reports closed-loop performance
across front-initialized and omnidirectional tracking tasks.
Direct-inference policies exhibit limited transferability
to panoramic UAV control, achieving zero episode-level
success without benchmark adaptation.
Relative to the fine-tuned baselines with PVR,
introducing CTM and GTS improves overall ATF from
229.93 to 375.29 frames with $\pi_0$ and from
244.21 to 337.21 frames with $\pi_{0.5}$,
corresponding to improvements of 63.2\% and 38.1\%,
respectively.
Normalized ATF also increases by 12.72 and 7.96
percentage points, indicating improvements in effective
tracking duration across variable episode horizons.
Both variants achieve an online SR of 10.67\%,
compared with 5.33\% and 6.67\% for their
fine-tuned counterparts.
Although full-episode completion remains challenging,
the consistent gains across both backbones demonstrate
the effectiveness of 360UAV-TVLA in prolonging
closed-loop target following.

\subsection{Offline Expert Replay}
\label{sec:exp_observation}

Table~\ref{tab:offline_expert_replay} shows that offline
imitation accuracy does not necessarily predict closed-loop
tracking performance.
Although 360UAV-TVLA improves online tracking duration,
its overall offline UAV ADE and distance-based SR
do not improve.
Expert replay measures trajectory agreement under
recorded observations, whereas closed-loop evaluation
accounts for action-dependent observations, accumulated
control errors, and inference delays.
These results highlight the necessity of evaluating
embodied tracking in closed loop alongside offline metrics.

\begin{table}[t]
\centering
\small
\caption{\textbf{Task-wise online success rates (\%).}
This table decomposes the online tracking performance across
the four subtasks of Task I (A1--A4) and Task II (B1--B4),
with their overall SR.}
\label{tab:taskwise_tracking}
\vspace{-0.9em}
\setlength{\tabcolsep}{3pt}
\renewcommand{\arraystretch}{1.0}

\resizebox{\linewidth}{!}{%
\begin{tabular}{@{}llllllllll@{}}
\toprule
& \multicolumn{4}{c}{Task I}
& \multicolumn{4}{c}{Task II}
& Overall \\
\cmidrule(lr){2-5}
\cmidrule(lr){6-9}
\cmidrule(l){10-10}
Method & A1 & A2 & A3 & A4
& B1 & B2 & B3 & B4 & (150) \\
\midrule

$\pi_0$ (w/ ft)
& 9.52 & \underline{0.00} & 0.00 & \underline{0.00}
& 6.82 & 0.00 & 0.00
& 8.33 & 5.33 \\

$\pi_{0.5}$ (w/ ft)
& \underline{14.29} & \underline{0.00} & 0.00 & \underline{0.00}
& \underline{9.09} & 0.00 & 0.00
& 8.33 & \underline{6.67} \\

\midrule

\rowcolor{blue!4}
\textbf{Ours} ($\pi_0,\lambda=2$)
& 4.76 & \textbf{28.57} & 0.00 & \textbf{5.56}
& \textbf{13.64} & 0.00 & \textbf{15.38}
& \underline{11.11} & \textbf{10.67} \\

\rowcolor{blue!4}
\textbf{Ours} ($\pi_{0.5},\lambda=2$)
& \textbf{19.05} & \underline{0.00} & 0.00 & \textbf{5.56}
& \underline{9.09} & 0.00 & \underline{7.69}
& \textbf{16.67} & \textbf{10.67} \\

\bottomrule
\end{tabular}%
}
\vspace{-1.3em}
\end{table}
\begin{table}[t]
\centering
\small
\caption{\textbf{Online tracking under different initial velocities.}
Results use the model-compute-delay protocol.
Groups are defined by the UAV's initial velocity;
ATF is reported in frames and SR in \%.}
\label{tab:online_initial_velocity}
\vspace{-1.0em}
\setlength{\tabcolsep}{4.5pt}
\renewcommand{\arraystretch}{0.95}

\resizebox{\linewidth}{!}{%
\begin{tabular}{@{}lllllll@{}}
\toprule
& \multicolumn{2}{c}{Zero Init. (75)}
& \multicolumn{2}{c}{Nonzero Init. (75)}
& \multicolumn{2}{c}{Overall (150)} \\
\cmidrule(lr){2-3}
\cmidrule(lr){4-5}
\cmidrule(l){6-7}
Method
& ATF$\uparrow$ & $\mathrm{SR}_{\mathrm{on}}\uparrow$
& ATF$\uparrow$ & $\mathrm{SR}_{\mathrm{on}}\uparrow$
& ATF$\uparrow$ & $\mathrm{SR}_{\mathrm{on}}\uparrow$ \\
\midrule

$\pi_0$ (w/o ft)
& 38.81 & 0.00
& 40.52 & 0.00
& 39.67 & 0.00 \\

$\pi_{0.5}$ (w/o ft)
& 49.00 & 0.00
& 48.17 & 0.00
& 48.59 & 0.00 \\

$\pi_0$ (w/ ft)
& 190.32 & 1.33
& 269.53 & \underline{9.33}
& 229.93 & 5.33 \\

$\pi_{0.5}$ (w/ ft)
& 219.23 & 4.00
& 269.19 & 9.33
& 244.21 & 6.67 \\

\midrule

\rowcolor{blue!4}
\textbf{Ours} ($\pi_0,\lambda=2$)
& \textbf{311.03} & \underline{5.33}
& \textbf{439.56} & \textbf{16.00}
& \textbf{375.29} & \textbf{10.67} \\

\rowcolor{blue!4}
\textbf{Ours} ($\pi_{0.5},\lambda=2$)
& \underline{307.40} & \textbf{8.00}
& \underline{367.01} & \underline{13.33}
& \underline{337.21} & \textbf{10.67} \\

\bottomrule
\end{tabular}%
}
\vspace{-0.5em}
\end{table}

\begin{figure}[t]
    \centering
    \includegraphics[width=\linewidth]{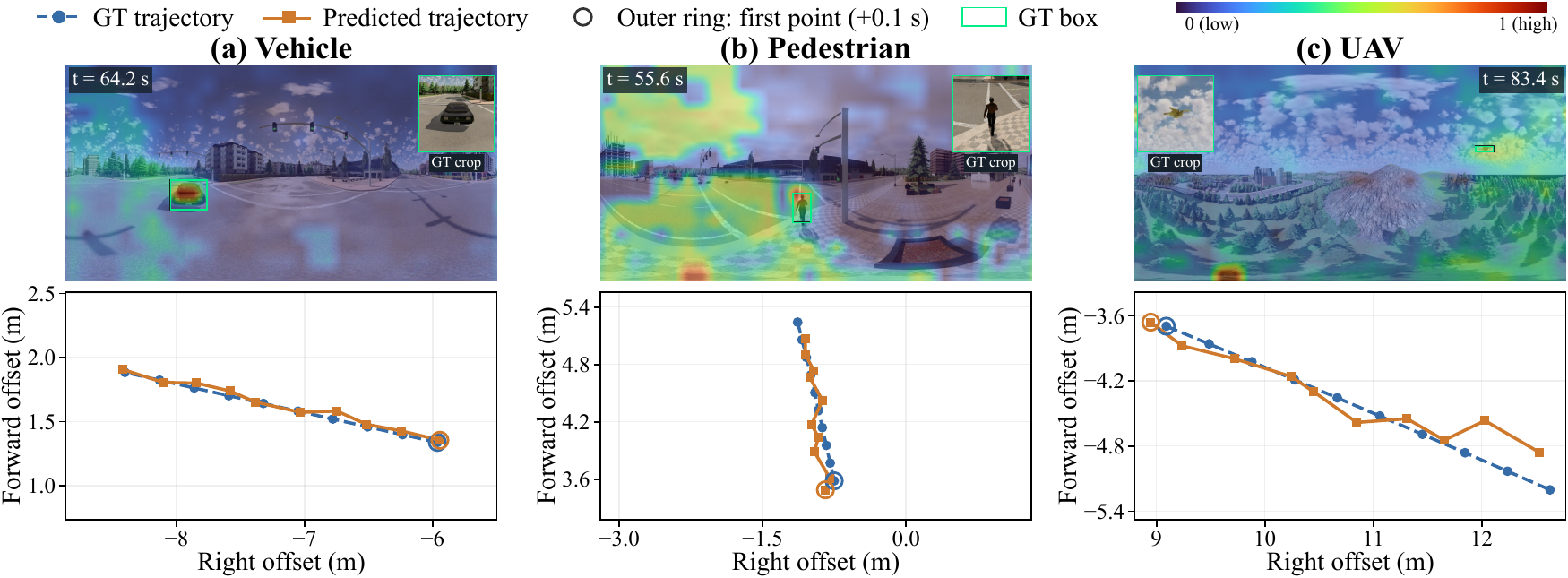}
    \vspace{-1.8em}
    \caption{\textbf{GTS trajectory prediction.}
    Horizontal projections of predicted and ground-truth
    target trajectories.}
    \label{fig:qual_target_future}
    \vspace{-0.6em}
\end{figure}

\subsection{Ablation of the Proposed Components}
\label{sec:exp_components}

Table~\ref{tab:component_ablation} evaluates the contributions
of PVR, CTM, and GTS.
PVR improves normalized ATF from 17.96\% to 25.68\%,
highlighting the importance of instance-specific conditioning.
Adding CTM or GTS to PVR further increases normalized
ATF to 28.16\% and 31.61\%, respectively.
Combining all three components achieves the highest ATF
of 337.21 frames and normalized ATF of 33.64\%.
Although PVR+GTS achieves a higher SR,
the full model provides the best tracking persistence.

\subsection{Analysis of Key Design Parameters}
\label{sec:exp_design}

\paragraph{CTM history length.}
Table~\ref{tab:online_ablation}(a) compares
$T\in\{0,1,2,4,8\}$ past frames with 128 tokens per frame.
$T=2$ achieves the highest ATF and normalized ATF,
337.21 frames and 33.64\%, whereas $T=0$ achieves
the highest SR of 12.00\%.
$T=8$ reaches 324.69 frames and 32.91\% normalized ATF,
but reduces inference throughput to 13.79 decisions/s.
These results reveal a trade-off among temporal context,
inference efficiency, and closed-loop tracking performance.
Among the evaluated settings, $T=2$ provides the
best tracking persistence while maintaining
18.90 decisions/s.

\paragraph{GTS loss weight.}
Table~\ref{tab:online_ablation}(b) evaluates different GTS
loss weights with a fixed history length of $T=2$.
Increasing the GTS loss weight does not consistently improve
performance. Normalized ATF increases from 26.96\% to 32.53\%
as the weight increases from 0.01 to 0.1, but drops to 27.87\%
at 0.5.
With $\lambda_{\mathrm{traj}}=2$, the model achieves the highest
ATF, normalized ATF, and SR of 337.21 frames, 33.64\%,
and 10.67\%, respectively.
Further increasing the weight to 5 reduces SR to 6.67\%.
These results highlight the importance of balancing
trajectory supervision with action imitation.

\subsection{Full-Spherical Observation vs. Active Steering}
\label{sec:erp_or_per}

Table~\ref{tab:erp_fixed_heading_ablation} compares three
observation--control configurations: fixed-heading perspective
perception, fixed-heading panoramic perception,
and perspective perception with active yaw control.
With the $\pi_{0.5}$ backbone, replacing perspective
observations with ERP improves ATF by 128.06 and
126.85 frames on Task I and Task II, respectively.
These results highlight the benefits of expanded angular
coverage for tracking persistence and non-frontal acquisition.
The evaluated active-steering policy underperforms
both fixed-heading configurations.
However, its yaw supervision is synthesized along
recorded translational trajectories rather than obtained
from jointly optimized translation--yaw demonstrations.
Thus, this comparison supports full-spherical perception
under the evaluated settings without establishing
an inherent disadvantage of active steering.

\begin{figure}[t]
    \centering
    \includegraphics[width=\linewidth]{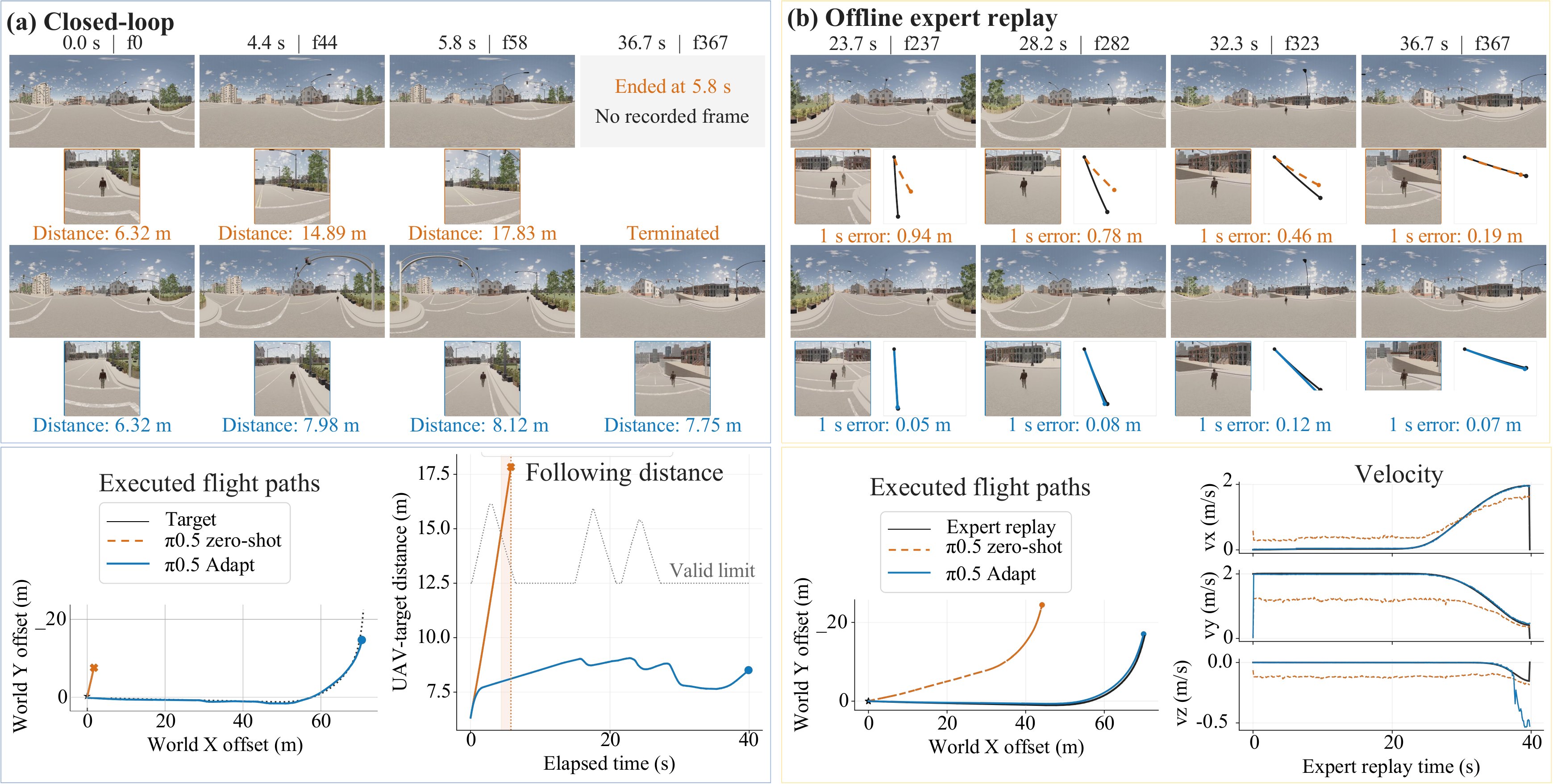}
    \vspace{-1.8em}
    \caption{\textbf{Effect of benchmark adaptation on $\pi_{0.5}$.}
    (a) Executed trajectories and target-following distances in
    closed loop. (b) Integrated trajectories and velocity
    predictions under expert replay.}
    \label{fig:qual_pi05}
    \vspace{-0.7em}
\end{figure}

\subsection{Task and Initialization Effects}

Table~\ref{tab:taskwise_tracking} reveals complementary
task-wise strengths: $\pi_0$ performs better on A2 and B3,
whereas $\pi_{0.5}$ leads on A1 and B4.
Table~\ref{tab:online_initial_velocity} compares 75 stationary
and 75 moving starts.
Relative to fine-tuned baselines, our $\pi_0$ variant improves
SR from 1.33\%/9.33\% to 5.33\%/16.00\%, while $\pi_{0.5}$
improves from 4.00\%/9.33\% to 8.00\%/13.33\%.
Both variants benefit from moving starts while consistently
improving performance across initialization regimes.
\textbf{Failure analysis.}
Both variants achieve 10.67\% overall SR, with no successes
on A3 or B2.
Test episodes average 1,166.7 frames, with a median of
1,200 frames (120\,s).
Success requires completing the entire horizon under visibility
and distance constraints without triggering the 15-frame
failure threshold, highlighting the difficulty of sustained
long-horizon tracking.

\subsection{Qualitative Visualization}
\label{sec:qualitative}

\paragraph{Future Target Trajectory Prediction.}
Figure~\ref{fig:qual_target_future} visualizes GTS predictions
over a one-second horizon for vehicle, pedestrian, and UAV targets.
Predicted and ground-truth trajectories share the current UAV origin
and fixed-heading control axes, exposing both localization offsets
and motion-direction differences in their horizontal projections.
Examples follow a fixed class-wise selection rule independent
of prediction error.
These predictions characterize learned target dynamics,
rather than the integrated UAV trajectory evaluated in expert replay.
The GTS predictions are inspected offline and are neither
executed as commands nor supplied to the action decoder.

\paragraph{Velocity Prediction under Expert Replay.}
Figure~\ref{fig:qual_pi05}(b) compares zero-shot and fine-tuned
$\pi_{0.5}$ under identical expert observations and motion states.
In the illustrated sequence, adaptation brings velocity predictions
and their integrated trajectory closer to the expert.
Since observations remain fixed, this comparison measures
action imitation without reflecting action-dependent
distribution shifts.
It demonstrates the effect of benchmark adaptation
rather than isolating individual contributions of
PVR, CTM, or GTS.

\begin{figure}[t]
    \centering
    \includegraphics[width=\linewidth]{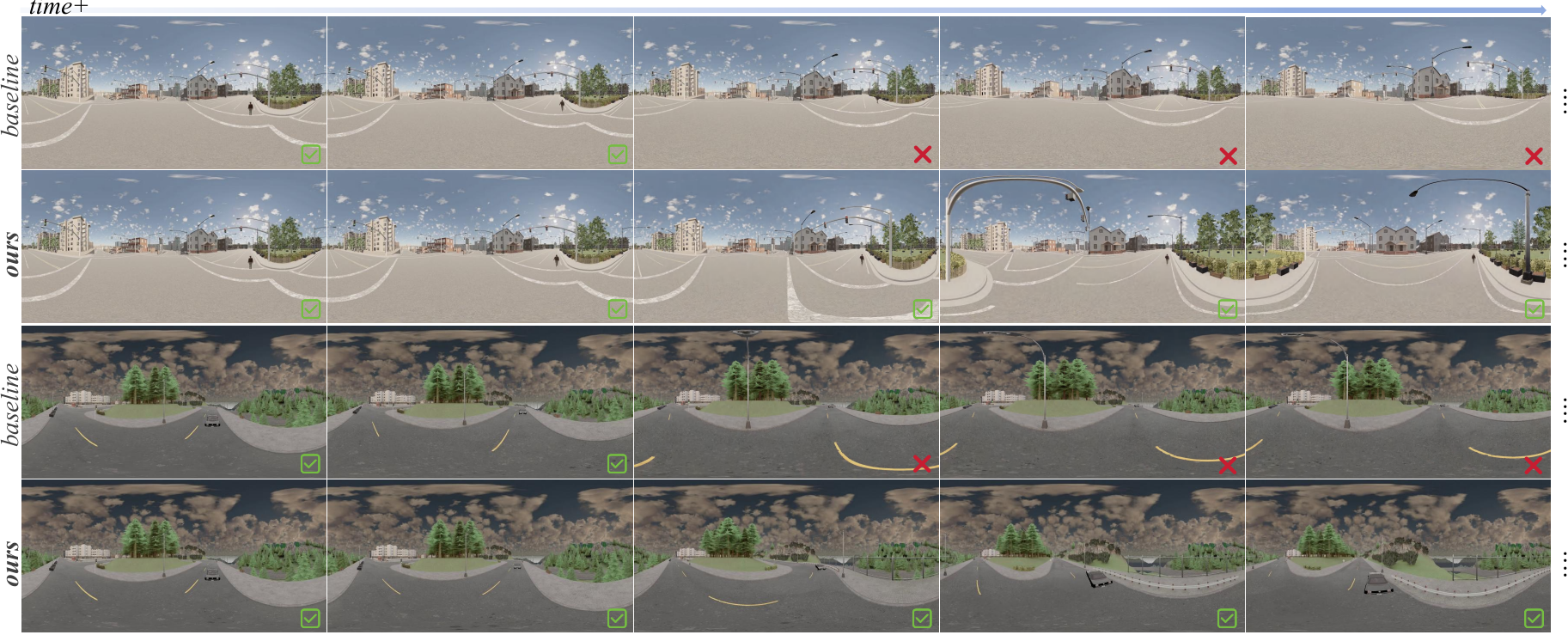}
    \vspace{-2.0em}
    \caption{\textbf{Qualitative closed-loop tracking comparison.}
    Panoramic observations at matched elapsed times for the
    fine-tuned baseline and 360UAV-TVLA under identical
    scenario initializations and the model-compute-delay protocol.}
    \label{fig:qual_tracking}
    \vspace{-1.0em}
\end{figure}

\paragraph{Closed-Loop Tracking Behavior.}
Figure~\ref{fig:qual_pi05}(a) shows the zero-shot policy terminating
at 5.8\,s, with a target distance of 17.83\,m versus
8.12\,m for the adapted policy.
Figure~\ref{fig:qual_tracking} compares matched panoramic
sequences, where 360UAV-TVLA maintains tracking after
the fine-tuned baseline loses it.
These examples illustrate improved tracking persistence
under action-dependent observations, complementing
the aggregate ATF, normalized ATF, and SR results.
\section{Conclusion}
\label{sec:conclusion}

We introduce \textbf{360UAV-ETrack}, a large-scale benchmark
for full-spherical embodied UAV tracking under fixed-heading control,
supporting both offline expert replay and online closed-loop evaluation.
To address the challenges of instance ambiguity, temporal motion
reasoning, and perception-to-control alignment, we propose
\textbf{360UAV-TVLA}, a panoramic VLA framework integrating
persistent visual referring, compressed temporal memory,
and geometry-aligned trajectory supervision.
Experiments across two VLA backbones demonstrate consistent
improvements in closed-loop tracking and reveal a gap between offline imitation and online
tracking.
These findings highlight the potential of full-spherical perception
for embodied UAV tracking without active heading adjustment.
\textbf{Future work.} We will explore lightweight models and real-world panoramic UAV
deployment to improve tracking efficiency and robustness.

\newpage
\section*{LLM Usage Statement}

Large language models were used to assist with language
polishing and manuscript formatting. The authors reviewed
and verified all resulting content and take full
responsibility for the final manuscript.

{
    \small
    \bibliographystyle{ieeenat_fullname}
    \bibliography{main}
}

\clearpage
\setcounter{page}{1}
\maketitlesupplementary
\appendix

\newpage
\appendix

\section{Detailed Online Evaluation Protocols}
\label{app:timing_protocols}
\mbox{}

All policy outputs are mapped to three-dimensional velocity
commands in the heading-aligned frame using fixed,
model-specific adapters.
Only the first command in each predicted action chunk
is executed. The three protocols differ in how inference
and external overhead affect simulated time:
\begin{equation}
D_{\mathrm{(i)}} = 0,
D_{\mathrm{(ii)}} = L_{\mathrm{model}} + L_{\mathrm{ext}},
D_{\mathrm{(iii)}} = L_{\mathrm{model}},
\end{equation}
where $L_{\mathrm{model}}$ is the measured model computation
latency and $L_{\mathrm{ext}}$ denotes external overhead,
including image preparation, communication, rendering,
and logging. These expressions exclude any additional
waiting caused by physics-step quantization.

\paragraph{Protocol (i): Synchronous Closed Loop.}
\begin{enumerate}
    \item At simulation time $t$, capture the image and
    measured UAV velocity, and start policy inference.
    \item Pause the simulator during inference.
    Neither the UAV nor the target advances in simulated time.
    \item Once inference completes, apply the first predicted
    velocity command and resume simulation.
    \item Hold this command for $0.1\,\mathrm{s}$ of simulated
    time, then capture the next observation and repeat.
\end{enumerate}
Thus, inference and external overhead affect experiment
wall time but introduce no delay in simulated time.

\paragraph{Protocol (ii): End-to-End Real-Time Closed Loop.}
\begin{enumerate}
    \item At simulation time $t$, capture the image and
    measured UAV velocity, and process the observation
    through the asynchronous inference pipeline.
    \item Keep the simulator running throughout observation
    preparation, communication, and inference.
    The UAV continues executing its previous velocity
    command while the target continues moving.
    \item When the prediction reaches the simulator,
    apply its first velocity command at the next available
    physics update.
    \item Hold the new command until the next prediction
    becomes effective, repeating the process with updated
    observations.
\end{enumerate}
This protocol exposes the policy to the full runtime delay,
including model computation and external overhead.

\paragraph{Protocol (iii): Model-Compute-Delay Closed Loop.}
\begin{enumerate}
    \item At simulation time $t$, capture the image and
    measured UAV velocity at the same timestamp, and start
    inference. Measure $L_{\mathrm{model}}$ from policy entry
    until the complete action chunk is available as a
    synchronized CPU array, including visual encoding
    and action generation.
    \item Advance simulated time over the measured inference
    interval. The UAV continues executing its previous
    velocity command while the target continues moving.
    \item Apply the first predicted command at the first
    $30\,\mathrm{Hz}$ physics boundary at or after
    $t+L_{\mathrm{model}}$:
    \begin{equation}
        t_{\mathrm{apply}}
        =
        \Delta_{\mathrm{phys}}
        \left\lceil
        \frac{t+L_{\mathrm{model}}}{\Delta_{\mathrm{phys}}}
        \right\rceil,
        \qquad
        \Delta_{\mathrm{phys}}=\frac{1}{30}\,\mathrm{s}.
    \end{equation}
    \item Hold the new command until the next prediction
    becomes effective. Only one prediction is logically
    pending at a time; subsequent predictions use the
    latest captured observation.
\end{enumerate}

Observations and scoring occur at $10\,\mathrm{Hz}$.
The causal history buffer receives every capture, including frames
acquired while a prediction is pending. Each request freezes its
three-frame window; the buffer is reset between episodes.
External image preparation, communication, rendering,
and logging affect wall time but are excluded from the
simulated delay.

We use protocol (iii) to approximate onboard execution
by retaining model computation latency while excluding
external overhead, particularly communication and rendering.
All online results in the main text follow protocol (iii).
Results under protocols (i) and (ii) are presented in this
appendix to assess performance without inference delay
and under full end-to-end runtime delay, respectively.

\section{Evaluation Metrics}
\label{app:evaluation_metrics}

We evaluate offline trajectory prediction, online closed-loop tracking,
and policy inference efficiency. All methods use fixed-heading
three-dimensional velocity control. The shared online evaluation setup
is described in Section~\ref{app:online_evaluation_protocol}, and the
main-text online results use timing protocol (iii) in
Section~\ref{app:timing_protocols}.
For episode $i$, let $H_i$ denote the prescribed number of observation
timestamps, indexed by $t=1,\ldots,H_i$, with
$\Delta t=0.1\,\mathrm{s}$. Unless stated otherwise, episode-level
metrics are averaged with equal weight over the $N$ evaluated episodes.

\subsection{Offline Average Displacement Error (ADE)}
\label{app:offline_ade}

Offline expert replay measures the trajectory implied by policy actions
under a fixed sequence of recorded expert observations.
At each timestamp, the policy receives the recorded expert observation
and motion-state input. We retain only the first predicted velocity
$\hat{\mathbf v}_{i,t}\in\mathbb R^3$ from its action chunk.
All subsequent inputs remain from the expert recording; reconstructed
positions do not affect the observations supplied to the policy.
The replay motion-state input is estimated by a causal backward
difference of recorded expert positions, with zero velocity at the
first timestamp. This is the offline replay input convention for the
historical expert recording; it is separate from the mixed tracker
initialization used in online Test episodes.

Let $\mathbf p^*_{i,t}$ be the actual recorded expert tracker position
and $R_i$ the fixed rotation from the heading-aligned body frame to
the world frame. Starting from the same initial position, we reconstruct
the complete predicted trajectory as
\begin{equation}
\begin{aligned}
\hat{\mathbf p}_{i,1}&=\mathbf p^*_{i,1},\\
\hat{\mathbf p}_{i,t}&=\mathbf p^*_{i,1}
  +\Delta t\sum_{k=1}^{t-1}R_i\hat{\mathbf v}_{i,k},
  \quad t=2,\ldots,H_i.
\end{aligned}
\label{eq:offline_trajectory_integration}
\end{equation}
The predicted velocity at timestamp $t$ determines motion toward
timestamp $t+1$. Thus, $H_i$ recorded positions provide reference
endpoints for $H_i-1$ actions; the last predicted action is not
integrated because its next reference position is unavailable.
We apply no post-hoc translation, rotation, scaling, or temporal
alignment to the reconstructed trajectory.

We define the full-trajectory ADE as
\begin{equation}
\mathrm{ADE}=\frac{1}{N}\sum_{i=1}^{N}\frac{1}{H_i}
\sum_{t=1}^{H_i}
\left\|\hat{\mathbf p}_{i,t}-\mathbf p^*_{i,t}\right\|_2.
\label{eq:offline_ade}
\end{equation}
ADE is reported in meters (m), with lower values indicating better
time-aligned trajectory agreement. Each episode is scored over its
full recorded horizon, including timestamps after any offline
distance-based failure. This metric measures accumulated position
error rather than a single-step relative translation error.
Because the reference consists of actual recorded positions, ADE
also reflects discrepancies between command integration and simulated
motion. Integrating the expert commands provides a diagnostic of this
execution discrepancy; repeating the supplied current velocity provides
a motion-state baseline.

\subsection{Online Success Rate (SR)}
\label{app:online_sr}

Let $\mathbf p^{\mathrm{on}}_{i,t}$ and
$\mathbf g^{\mathrm{on}}_{i,t}$ be the online tracker and designated
target positions. Their three-dimensional distance is
$d^{\mathrm{on}}_{i,t}=
\|\mathbf p^{\mathrm{on}}_{i,t}-\mathbf g^{\mathrm{on}}_{i,t}\|_2$.
Let $z_{i,t}\in\{0,1\}$ indicate target visibility as assessed by the
evaluator. Using the scenario-specific lower bound $d_i^{\min}$ and
the potentially frame-dependent upper bound $d_{i,t}^{\max}$ defined
in Section~\ref{sec:tracking_distance_constraints}, a valid frame is
\begin{equation}
q^{\mathrm{on}}_{i,t}=\mathbb I\!\left[
z_{i,t}=1\ \land\
d_i^{\min}\leq d^{\mathrm{on}}_{i,t}\leq d_{i,t}^{\max}
\right].
\label{eq:online_valid_frame}
\end{equation}
The consecutive-invalid-frame counter is
\begin{equation}
c^{\mathrm{on}}_{i,t}=\begin{cases}
0,&q^{\mathrm{on}}_{i,t}=1,\\
c^{\mathrm{on}}_{i,t-1}+1,&q^{\mathrm{on}}_{i,t}=0,
\end{cases}
\quad c^{\mathrm{on}}_{i,0}=0.
\label{eq:online_invalid_streak}
\end{equation}
An episode terminates at the first timestamp when
$c^{\mathrm{on}}_{i,t}\geq15$. Invalid streaks shorter than 15 frames
are tolerated, and a valid frame resets the counter.
The online success indicator equals one only when the episode reaches
all $H_i$ prescribed timestamps without triggering this failure rule:
\begin{equation}
\begin{aligned}
s^{\mathrm{on}}_i&=\mathbb I\!\left[
  \text{episode reaches }H_i\text{ without failure}\right],\\
\mathrm{SR}_{\mathrm{on}}&=\frac{100}{N}\sum_{i=1}^N s^{\mathrm{on}}_i.
\end{aligned}
\label{eq:online_sr}
\end{equation}
SR is reported as a percentage (\%), with higher values indicating
more reliable completion of the prescribed tracking horizons.
SR and ATF evaluate target-following performance under
visibility and distance constraints.

\subsection{Online Average Tracking Frames (ATF)}
\label{app:online_atf}

Let $\tau_i$ be the first failure timestamp, or $H_i$ if the episode
completes without failure. The number of valid tracking frames and
its episode average are
\begin{equation}
L_i=\sum_{t=1}^{\tau_i}q^{\mathrm{on}}_{i,t},\qquad
\mathrm{ATF}=\frac{1}{N}\sum_{i=1}^N L_i.
\label{eq:online_atf}
\end{equation}
ATF is reported in observation frames, with higher values indicating
longer effective tracking duration. It counts valid frames rather than
all elapsed frames or simulator ticks, and therefore reflects partial
tracking performance even for unsuccessful episodes. Prescribed
horizons differ across episodes as summarized in
Table~\ref{tab:all_online_tasks}.

\subsection{Offline Distance Success Rate}
\label{app:offline_distance_sr}

For the reconstructed trajectory in
Eq.~\eqref{eq:offline_trajectory_integration}, let
$\mathbf g^*_{i,t}$ denote the recorded target position at the same
timestamp. Define
\begin{equation}
\begin{aligned}
d^{\mathrm{off}}_{i,t}&=
\left\|\hat{\mathbf p}_{i,t}-\mathbf g^*_{i,t}\right\|_2,\\
q^{\mathrm{off}}_{i,t}&=\mathbb I\!\left[
d_i^{\min}\leq d^{\mathrm{off}}_{i,t}\leq d_{i,t}^{\max}
\right].
\end{aligned}
\label{eq:offline_distance_valid_frame}
\end{equation}
Using the same counter recurrence as
Eq.~\eqref{eq:online_invalid_streak}, with
$q^{\mathrm{off}}_{i,t}$ replacing $q^{\mathrm{on}}_{i,t}$ and
$c^{\mathrm{off}}_{i,0}=0$, we define
\begin{equation}
\begin{aligned}
s^{\mathrm{off,dist}}_i&=\mathbb I\!\left[
\max_{1\leq t\leq H_i}c^{\mathrm{off}}_{i,t}<15\right],\\
\mathrm{SR}_{\mathrm{off,dist}}&=
\frac{100}{N}\sum_{i=1}^N s^{\mathrm{off,dist}}_i.
\end{aligned}
\label{eq:offline_distance_sr}
\end{equation}
The original distance bounds and 15-frame threshold are shared with
online evaluation. However, this offline metric does not assess
visibility, occlusion, or collision at reconstructed positions, and
recorded expert visibility is not treated as visibility from the
predicted trajectory. It is reported separately from online SR.
Replay and full-horizon ADE calculation continue after an offline
failure, while that episode's offline success indicator remains zero.

\subsection{Inference Speed (FPS)}
\label{app:inference_fps}

Policy inference throughput is defined as
\begin{equation}
\mathrm{FPS}=\frac{N_{\mathrm{dec}}}
{\sum_{j=1}^{N_{\mathrm{dec}}}T_j},
\label{eq:inference_fps}
\end{equation}
where $N_{\mathrm{dec}}$ is the number of evaluated decisions and $T_j$
is the model inference time for decision $j$. The timing boundary
includes visual encoding, multimodal processing, and action generation,
but excludes simulator execution and rendering. GPU execution must
complete before a timing interval is closed. Cross-method throughput
comparisons require matched hardware, batch size, warm-up treatment,
and timing boundaries. The 10\,Hz observation and scoring frequency and
the throughput of concurrently running replay jobs are not standalone
policy FPS measurements. Higher FPS indicates faster inference.

\section{Tracking Distance Constraints}
\label{sec:tracking_distance_constraints}

Table~\ref{tab:online_tracking_distance} lists the base intervals used
in the distance-validity tests of
Eqs.~\eqref{eq:online_valid_frame} and
\eqref{eq:offline_distance_valid_frame}. The lower bound $d_i^{\min}$
is fixed by target type. The upper bound $d_{i,t}^{\max}$ is the
precomputed per-frame value when a distance profile is specified;
otherwise, it equals the base upper bound. Both endpoints are inclusive.
These are evaluation thresholds rather than measured trajectory
distance coverage or maximum tracker operating distances.

Some ground-target scenarios use precomputed upper-bound profiles to
accommodate terrain-clearance adjustments to the reference flight
trajectory. Across the complete set, 96 episodes contain such profiles;
the upper bound exceeds its base value at some timestamps in 29 of
the 72 pedestrian episodes and 10 of the 72 vehicle episodes.
The largest individual upper bounds are 28.13\,m and 24.54\,m,
respectively. These values are not constant limits applied to all
frames or episodes. Lower bounds remain unchanged, and all six UAV/B2
episodes retain the fixed interval 2.5--15\,m.
All bounds are fixed before policy evaluation and shared across methods.
The six UAV/B2 episodes provide more limited coverage for this target
category than the pedestrian and vehicle episodes.

\begin{table}[t]
\centering
\caption{Base distance constraints for online tracking and offline
distance SR. Some ground-target episodes use precomputed per-frame
upper bounds as described in the text.}
\label{tab:online_tracking_distance}
\small
\begin{tabular}{lcc}
\toprule
Target Type & No. of Episodes & Base Interval (m) \\
\midrule
Pedestrian & 72 & 1.75--12.5 \\
Vehicle & 72 & 2.8--20 \\
UAV (B2) & 6 & 2.5--15 \\
\bottomrule
\end{tabular}
\end{table}

\section{Online Evaluation Protocol}
\label{app:online_evaluation_protocol}

\subsection{Task Composition}

The evaluation contains 150 episodes: 51 in Task I
(front-initialized persistent tracking; A1--A4) and 99 in Task II
(omnidirectional acquisition and tracking; B1--B4).
The set includes 72 pedestrian, 72 vehicle, and six UAV episodes,
with 50 episodes in each of Town01, Town03, and Town04.
Table~\ref{tab:all_online_tasks} gives the task composition and
prescribed horizons. Canonical task categories are used for aggregation;
retained legacy scenario identifiers do not determine the reported
category. Each episode has its own fixed horizon, and every method is
evaluated using the same episode definitions and thresholds.

\begin{table}[t]
\centering
\caption{Composition and prescribed horizons of all evaluation tasks.
Horizon ranges summarize assigned episode durations. Each individual episode has a fixed horizon.}
\label{tab:all_online_tasks}
\small
\setlength{\tabcolsep}{4pt}
\resizebox{1.0\linewidth}{!}{%
\begin{tabular}{llrrrrc}
\toprule
Task & Category & Total & Pedestrian & Vehicle & UAV & Horizon (s) \\
\midrule
A1 & Front tracking & 21 & 9 & 12 & 0 & 40--200 \\
A2 & Front underpass & 7 & 2 & 5 & 0 & 120--180 \\
A3 & Front approach & 5 & 3 & 2 & 0 & 160--200 \\
A4 & Front target selection & 18 & 8 & 10 & 0 & 40--200 \\
\midrule
Task I & Subtotal & 51 & 22 & 29 & 0 & 40--200 \\
\midrule
B1 & Non-front tracking & 44 & 25 & 19 & 0 & 40--200 \\
B2 & UAV tracking & 6 & 0 & 0 & 6 & 100 \\
B3 & Non-front approach & 13 & 7 & 6 & 0 & 40--200 \\
B4 & Non-front target selection & 36 & 18 & 18 & 0 & 40--200 \\
\midrule
Task II & Subtotal & 99 & 50 & 43 & 6 & 40--200 \\
All & Total & 150 & 72 & 72 & 6 & 40--200 \\
\bottomrule
\end{tabular}%
}
\end{table}

\subsection{Observations and Target Initialization}

All methods receive full-spherical equirectangular observations from
1920$\times$960 source images; internal preprocessing follows each
method's specified adapter. The designated target is identified once
in the actual first observation through a bounding-box annotation,
a language instruction, and a persistent reference image where
supported by the adapter. This initialization remains fixed throughout
the episode; subsequent target ground truth is not supplied as
per-frame annotations. The online motion-state input contains the measured tracker velocity.
The Test protocol assigns 75 episodes to zero-speed tracker
initialization and 75 episodes to nonzero-speed initialization.
For each nonzero initialization, the prescribed world-coordinate
tracker velocity equals the target motion reference vector specified
by the scenario controller. This reference vector need not equal the
target's measured velocity at frame zero during settling or acceleration.
The policy receives the tracker velocity measured at its actual first
observation, which may differ slightly from the prescribed value after
the physical initialization steps.

The nominal scenario initialization specifies target bearings of
$-45^\circ$, $0^\circ$, or $45^\circ$ for front-initialized episodes,
relative to the fixed tracker heading. Ground-target Task II scenarios
specify $-135^\circ$, $-90^\circ$, $90^\circ$, $135^\circ$, or
$180^\circ$; B2 scenarios specify $-155^\circ$ or $115^\circ$.
These are scenario initialization parameters; azimuths measured from
recorded observations are computed from the realized tracker and
target poses. A4 and B4 contain one or two explicitly
added distractors, whereas the other categories do not.
Distance and visibility are evaluated against the designated target,
not the nearest distractor.

\subsection{Control and Synchronization}

Policies output three-dimensional velocities in the fixed
heading-aligned body frame, with zero yaw rate. Observations and
scoring occur at 10\,Hz, while the simulator advances at 30\,Hz.
Only the first action of each predicted chunk is executed.
Commands whose three-dimensional speed exceeds 6\,m/s are rescaled
to this norm while preserving their direction.

Command timing follows the protocol in
Section~\ref{app:timing_protocols}. Under synchronous protocol (i),
the simulator pauses during inference, and each new command is held
for three physics ticks (0.1\,s). Under protocol (ii), the previous
command remains active during the full end-to-end runtime delay.
Main-text online results use protocol (iii): simulated time advances
over the measured model computation interval, while external overhead
is excluded from simulated delay. The previous command remains active
until the new prediction takes effect at the first eligible physics
boundary. Thus, protocols (ii) and (iii) do not generally have a fixed
three-tick command hold.

\subsection{Visibility, Termination, and Reporting}

Target visibility is assessed using simulator occlusion queries toward
multiple points on the target. At least one unobstructed probe is
sufficient for visibility. This geometric evaluator signal is used
only for scoring and is not a model detection-confidence threshold.
Together with the original distance bounds in
Section~\ref{sec:tracking_distance_constraints}, it determines online
frame validity under Eq.~\eqref{eq:online_valid_frame}.

Online episodes stop at the first 15-frame invalid streak or at their
prescribed horizon, as defined in Section~\ref{app:online_sr}.
At 10\,Hz, the failure tolerance is a nominal 1.5\,s window.
Runs interrupted by infrastructure errors are rerun to complete the
evaluation; such interruptions are not reported as completed successful
episodes. SR and ATF are aggregated separately for A1--A4, B1--B4,
Task I, Task II, and the full set. The complete evaluation uses
denominators of 51, 99, and 150 episodes for Task I, Task II, and
overall SR, respectively.

Offline expert replay is evaluated separately using ADE and offline
distance SR from Sections~\ref{app:offline_ade} and
\ref{app:offline_distance_sr}. Its fixed observations are not generated
by the reconstructed policy trajectory, so offline distance success
does not establish online closed-loop tracking success.

\section{Implementation Details of Compared Methods}
\label{app:baseline_implementation}

All methods are evaluated on the same 150 episodes under the
model-compute-delay protocol.
At each decision, the evaluator executes the first predicted
action and holds the previous velocity while model computation
advances the simulation clock.
Translational commands are expressed in the original
initial-heading coordinate frame and are bounded by a
maximum speed of $6\,\mathrm{m/s}$.
FPS is computed as the number of completed inference requests
divided by their total synchronized model-computation time,
including requests whose outputs are discarded upon episode
termination.
Model loading, warm-up, external image preparation, communication,
rendering, and logging are excluded.
For benchmark-trained models, checkpoint selection uses the
held-out validation split rather than online test performance.
No method receives current ground-truth target states during
deployment.

\paragraph{Uni-NaVid.}
We use the released Uni-NaVid-7B checkpoint~\citep{zhang2025uninavid}
without benchmark fine-tuning or additional trainable modules.
The complete ERP image is letterboxed to the native
$224\times224$ input resolution.
Its causal video context is sampled at 1\,Hz and contains at
most 32 frames, combining recent observations with uniformly
sampled earlier frames.
The native language interface predicts discrete navigation
commands using deterministic decoding.
A fixed adapter maps forward motion to a longitudinal velocity
of $0.5\,\mathrm{m/s}$.
Left, right, and stop commands produce zero translation because
the shared evaluation protocol fixes the UAV heading.
These commands are not reinterpreted as lateral motion.
Consequently, this baseline evaluates the released navigation
policy through a fixed interface with restricted UAV control
expressiveness.

\paragraph{OmTrackVLA.}
We deploy the released OmTrackVLA-0.6B
checkpoint~\citep{omtrackvla2025} without benchmark fine-tuning.
The policy receives the language instruction and the latest
32 causal ERP observations, each letterboxed to
$384\times384$.
Missing initial history is padded by repeating the earliest
available observation.
The native planner predicts eight planar waypoints with
associated yaw values.
We convert successive planar waypoint displacements into
horizontal velocities using a $0.1$\,s interval, set vertical
velocity to zero, and discard yaw commands under the
fixed-heading protocol.
The resulting commands are passed through the common action
executor.
No learned UAV action head, target-state input, or benchmark
adaptation is introduced.

\paragraph{$\pi_0$ without fine-tuning.}
We evaluate the published OpenPI $\pi_0$ base
checkpoint~\citep{black2025pi0} using a parameter-free mapping
to its existing observation and action interfaces.
The current ERP image is letterboxed to $224\times224$ and
assigned to the main camera slot.
A static target-context crop from the first ERP observation
occupies an existing secondary camera slot, while the unused
camera slot is masked.
The normalized UAV velocity fills the first three native
state coordinates, with the remaining coordinates set to zero.
The first three continuous action channels are interpreted
as XYZ velocities and converted to physical units using the
benchmark training-set normalization.
Native ten-step flow sampling is retained, and only the first
predicted velocity is executed.
No policy parameters are updated.

\paragraph{$\pi_{0.5}$ without fine-tuning.}
The direct-inference $\pi_{0.5}$ baseline uses the published
OpenPI checkpoint~\citep{black2025pi05} and the same fixed
camera-slot and velocity-channel mapping as the direct
$\pi_0$ baseline.
The current ERP image, initial target-context reference,
and language instruction are supplied through existing
input interfaces.
The UAV velocity is represented using the native
$\pi_{0.5}$ state-conditioning procedure.
We retain the original ten-step flow sampler and map the
first three predicted action channels to physical XYZ
velocities using the shared training-set statistics.
No benchmark fine-tuning, CTM, GTS, or learned
action calibration is applied.

\paragraph{Temporal BC.}
Temporal BC is trained from scratch on the benchmark
demonstrations.
A shared convolutional encoder processes the current ERP
observation, two preceding observations at
$[t-0.2,t-0.1]$, and the static target reference.
A 256-dimensional gated recurrent unit aggregates the
causal image history, with invalid historical observations
masked.
The temporal representation is concatenated with reference
features, encoded UAV velocity, and a target-category code
derived from the instruction.
A feed-forward decoder predicts ten future XYZ velocities.
Training minimizes masked velocity MAE in physical units
using AdamW with learning rate $3\times10^{-4}$,
weight decay $0.01$, global batch size 64, and seed 42
for 21,091 updates.
The checkpoint with the lowest held-out validation error
is used for closed-loop evaluation.

\paragraph{D-VAT-BC.}
D-VAT-BC is a supervised adaptation of the D-VAT actor
architecture~\citep{dionigi2024dvat}, rather than an evaluation
of its original reinforcement-learning policy.
An ImageNet-pretrained ResNet-18 encodes the three causal
ERP observations and the static target reference.
Historical features are concatenated and projected before
fusion with reference features, UAV velocity, and the
instruction-derived target-category code.
The actor-style multilayer perceptron is retained, while
the original control output is replaced with a decoder
for ten future XYZ velocities.
The model is trained with the same masked physical-velocity
MAE, optimizer settings, sample budget, and validation
selection procedure as Temporal BC.
No reinforcement-learning reward, additional environment
interaction, or online target-state supervision is used.

\paragraph{Fast-WAM.}
We initialize Fast-WAM~\citep{yuan2026fastwam} from its released
two-camera LIBERO checkpoint and fine-tune it on the
benchmark demonstrations.
The current ERP observation and static target reference
are encoded through its image-latent pathway, while the
language instruction and normalized UAV velocity provide
additional conditioning.
The action branch predicts a ten-step XYZ velocity chunk
using its native iterative sampler with ten sampling steps.
Benchmark adaptation minimizes the masked physical-velocity
MAE of the sampled actions; it does not introduce future
images as policy inputs or an additional benchmark video
prediction objective.
Training uses one epoch, global batch size 64, peak learning
rate $2.5\times10^{-5}$, and seed 42.
The selected policy executes only its first predicted action.

\paragraph{$\pi_0$ with fine-tuning.}
The fine-tuned $\pi_0$ baseline is initialized independently
from the published pretrained backbone~\citep{black2025pi0}.
It receives the current $448\times224$ ERP observation,
a $224\times224$ persistent target reference as in PVR,
the language instruction, and normalized UAV velocity.
The ERP aspect ratio is preserved by interpolating the
visual positional embeddings to the rectangular token grid.
The action interface predicts ten future three-dimensional
velocities.
We optimize the masked physical-velocity MAE through the
ten-step action sampler for one epoch, using AdamW,
global batch size 64, peak learning rate
$2.5\times10^{-5}$, and seed 42.
This baseline retains PVR but incorporates neither CTM nor GTS.

\paragraph{$\pi_{0.5}$ with fine-tuning.}
We fine-tune the published $\pi_{0.5}$
backbone~\citep{black2025pi05} using the same current ERP,
persistent target reference, instruction, and motion-state
inputs as the fine-tuned $\pi_0$ baseline.
The rectangular ERP representation and ten-step XYZ
prediction interface are shared, while the native
$\pi_{0.5}$ conditioning mechanism is retained.
Training uses the same scene-disjoint data split,
normalization, masked physical-velocity objective,
one-epoch update budget, and optimizer schedule.
This baseline retains PVR without CTM or GTS.
The validation-selected checkpoint is evaluated with
native ten-step flow sampling, executing only the first
predicted velocity.

\paragraph{360UAV-TVLA with $\pi_0$.}
Our $\pi_0$ variant is initialized independently from the
pretrained backbone and integrates persistent visual
referring (PVR), compressed temporal memory (CTM),
and geometry-aligned trajectory supervision (GTS).
PVR retains an instance-specific visual reference
throughout the episode.
CTM compresses each of the two preceding ERP observations
to 128 tokens, while preserving the full representations
of the current panorama and target reference.
Temporal offsets and observation-source embeddings
distinguish the visual inputs.
GTS predicts ten future target positions in the
current-origin, fixed-heading control frame during training.
The normalized trajectory loss is jointly optimized with
the velocity imitation objective under the common
one-epoch VLA training schedule.
During deployment, the GTS head is bypassed,
introducing no additional inference cost.

\paragraph{360UAV-TVLA with $\pi_{0.5}$.}
Our $\pi_{0.5}$ variant integrates the same PVR, CTM,
and GTS components into the pretrained $\pi_{0.5}$ backbone.
The current ERP observation contributes 512 visual tokens,
PVR contributes 256 reference tokens, and CTM contributes
128 tokens per historical frame, yielding 1,024 visual tokens
with two preceding frames.
Training jointly optimizes masked XYZ velocity MAE and
the GTS trajectory loss under the configured loss weight,
using the same data split, training budget, and
validation-based checkpoint selection.
During deployment, CTM maintains a causal 10\,Hz history
cache, including during simulated computation delays.
The policy executes only the first velocity generated
by native ten-step flow sampling, with the GTS head bypassed.

\section{Expert Demonstration Generation}
\label{app:expert_generation}

\subsection{Reference Trajectory Generation}

Expert demonstrations are generated using task-specific controllers
in CARLA. Pedestrians and vehicles follow navigable routes, while
aerial targets use dedicated three-dimensional motion controllers.
The UAV reference position is constructed from the target's
bounding-box center, its observed motion, and the desired following
geometry. Tracking sequences use target-relative offsets or
distance-based interpolation along the recorded target trail.
Where required, a smooth transition connects the initial offset
to the trailing reference.
Approach and acquisition tasks additionally include task-specific
initialization phases before entering sustained following.
In multi-target scenes, the reference remains associated with the
designated target.

\subsection{Velocity Control}

Controllers operate at 30\,Hz.
Let $\mathbf{p}_t^\star$ denote the reference position and
$\mathbf{p}_t$ the measured UAV position.
A feedforward--feedback controller produces
\begin{equation}
    \widetilde{\mathbf{v}}_t
    =
    \widehat{\mathbf{v}}_t^\star
    + k_p(\mathbf{p}_t^\star-\mathbf{p}_t),
\end{equation}
where $\widehat{\mathbf{v}}_t^\star$ is the filtered reference
velocity and $k_p$ is a task-specific position gain.
The command is constrained by speed and acceleration limits:
\begin{equation}
\begin{aligned}
    \overline{\mathbf{v}}_t
    &= \operatorname{sat}_{v_{\max}}
       (\widetilde{\mathbf{v}}_t),\\
    \mathbf{v}_t
    &= \mathbf{v}_{t-1}
       + \operatorname{sat}_{a_{\max}\Delta t}
       (\overline{\mathbf{v}}_t-\mathbf{v}_{t-1}),
\end{aligned}
\end{equation}
with $\Delta t=1/30$\,s and
$\operatorname{sat}_{b}(\mathbf{x})
=\mathbf{x}\min(1,b/\|\mathbf{x}\|_2)$,
defined as zero when $\mathbf{x}=\mathbf{0}$.
Gains and limits follow the corresponding task controller.
Commands are applied through CARLA's velocity-control interface.
Recorded action labels use the initial-heading coordinate frame
with an upward-positive vertical axis.

\subsection{Terrain Clearance and Obstacle Handling}

Initialization checks exclude UAV poses embedded in scene geometry
and select task-appropriate vertical offsets.
During motion, local geometry queries inspect the commanded
flight corridor, including lateral or vertical clearance margins.
Obstacle handling depends on the task: controllers may brake,
try locally checked climbing or lateral commands, or reject
the trial when the prescribed following geometry cannot be
maintained safely.
Collision callbacks and additional scene-specific geometry checks
identify invalid segments.
Only segments satisfying the corresponding collection and
quality-control checks are retained.
These procedures provide local safeguards rather than a
globally optimal collision-avoidance planner.

\subsection{Demonstration Recording}

The simulator advances synchronously at 30\,Hz, with panoramic
observations, UAV motion states, target annotations, and expert
velocity commands recorded every three physics ticks at 10\,Hz.
Sensor frames and actor states are matched by simulator frame
identifiers and timestamps.
Issued commands are stored separately from measured UAV
velocities.
The UAV maintains its initial heading throughout each
demonstration, with zero commanded yaw rate.
Simulator target states support expert generation and annotation;
target trajectories are not supplied to the deployed velocity policy.

\section{Quantitative Analysis of the Dataset}

Unless stated otherwise, trajectory-based statistics in this section
are computed from recorded expert sequences. Test tracker initialization
speeds are instead taken from the mixed zero/nonzero online evaluation
protocol. Historical expert states and protocol-prescribed online
initialization values are therefore distinguished throughout the analysis.

\subsection{Speed Distribution}

For each actor, let $\mathbf{p}_i$ denote the 3D position at timestamp $t_i$. The frame-wise speed is computed as

\begin{equation}
v_i =
\frac{\left\|\mathbf{p}_{i+1}-\mathbf{p}_i\right\|_2}
{t_{i+1}-t_i}.
\end{equation}

The average speed of one sequence is defined as its total traveled distance divided by its duration:

\begin{equation}
\bar{v} =
\frac{\sum_{i=0}^{T-2}
\left\|\mathbf{p}_{i+1}-\mathbf{p}_i\right\|_2}
{t_{T-1}-t_0}.
\end{equation}

This definition uses the synchronized 3D trajectory positions and actual timestamps. It includes stops and turning motion, and each sequence contributes one
equally weighted average-speed sample.

\begin{figure*}[t]
  \centering
  \includegraphics[width=\linewidth]{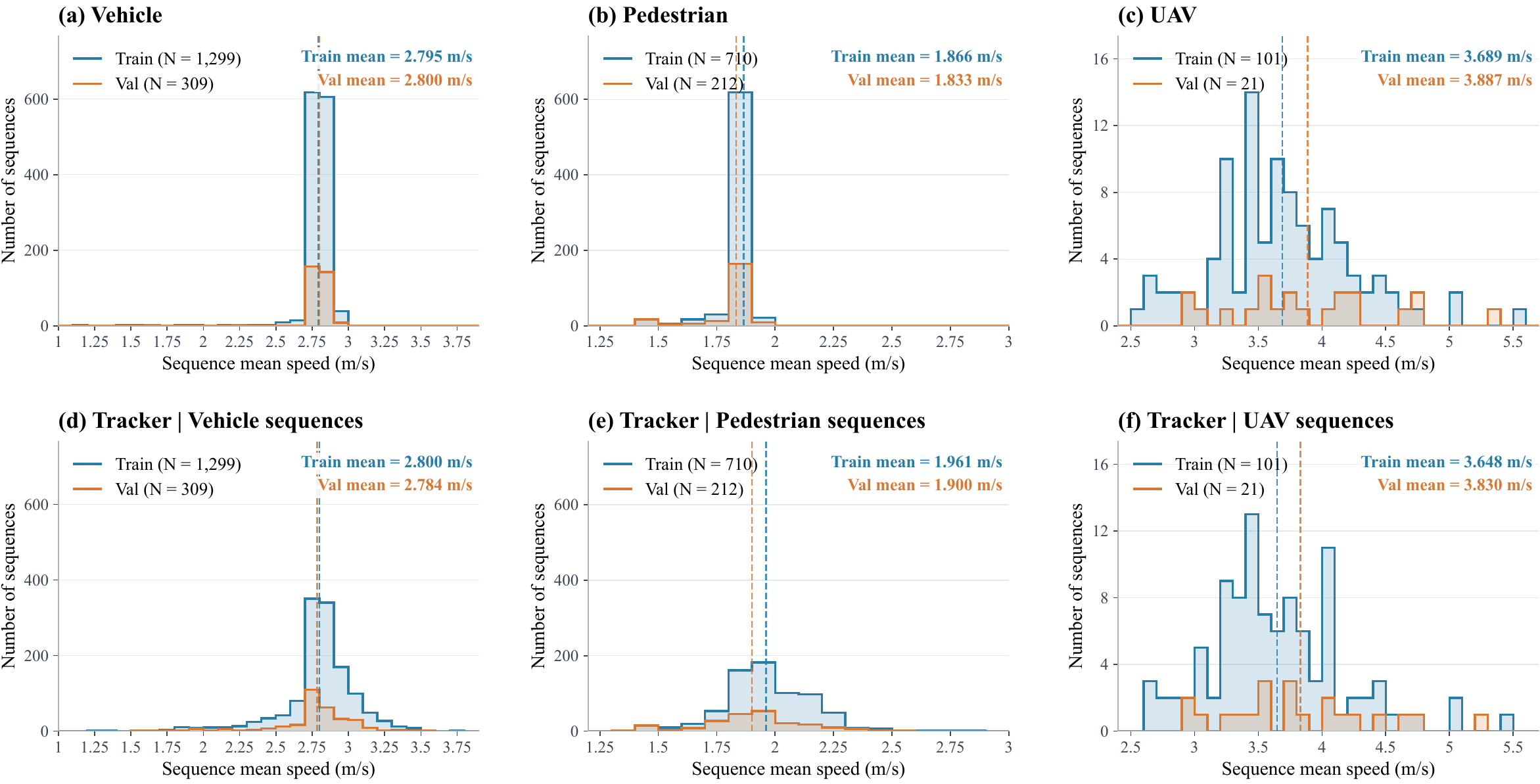}
  \caption{Sequence-average speed distributions for matched Target and Tracker sequences. The top row shows Target speeds and the bottom row shows the Tracker speeds for the same episodes. Columns correspond to Vehicle, Pedestrian, and UAV targets. Each sequence contributes exactly one observation, computed as cumulative three-dimensional displacement divided by sequence duration. Blue and orange dashed lines and labels denote the Train and Val means, respectively; the vertical axis reports the number of sequences.}
  \label{fig:panouav_appendix_sequence_speed}
\end{figure*}

Figure~\ref{fig:panouav_appendix_sequence_speed} shows three distinct target-speed regimes. Vehicle targets have average speeds of $2.795$ and $2.800$~m/s on
Train and Val, respectively, while pedestrian targets are slower at $1.866$ and $1.833$~m/s. UAV targets are faster and more variable, with averages of
$3.689$ and $3.887$~m/s. Tracker speeds follow the same class-dependent ordering, although they are slightly higher than the target speeds for pedestrian
sequences. The matched Target--Tracker distributions therefore indicate that the tracker motion is broadly consistent with the target-speed regime, while UAV
sequences provide the largest speed variation.
  
\begin{figure}[t]
    \centering
    \includegraphics[width=\linewidth]{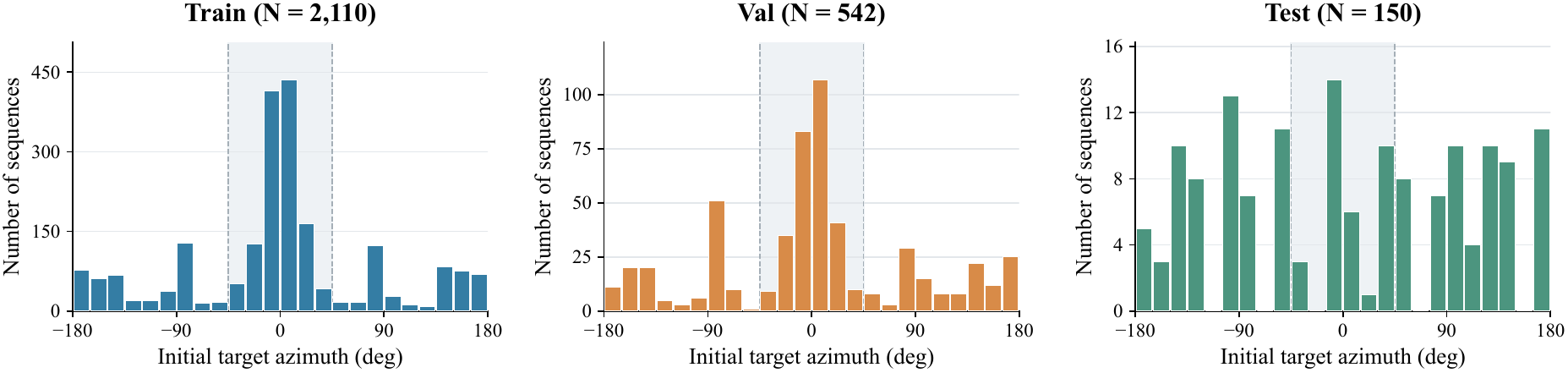}
    \caption{
        Initial target azimuth distributions across the Train, Val,
        and Test splits.
        Each sequence contributes one observation at its first recorded
        expert frame, with azimuth measured relative to the fixed tracker heading.
        Here, $0^\circ$ denotes the forward direction and
        $\pm180^\circ$ denotes the rear direction.
        All panels use $15^\circ$ bins, and the shaded region indicates
        the frontal horizontal $90^\circ$ field of view.
    }
    \label{fig:initial_azimuth}
\end{figure}

\subsection{Initial target azimuth coverage.}
Figure~\ref{fig:initial_azimuth} characterizes the target direction
in the first recorded expert observation of each episode.
Train and Val exhibit a similar concentration around the forward
direction, while also covering lateral and rear configurations.
Test is less concentrated near $0^\circ$, with initial target
directions distributed broadly across the full azimuth range.
These non-frontal initializations require the tracker to acquire
the designated target beyond a forward-facing camera's horizontal
$90^\circ$ field of view.
The distributions therefore motivate panoramic perception for
target acquisition under fixed-heading control, in addition to
maintaining observation as the target moves during tracking.

\begin{figure}[t]
    \centering
    \includegraphics[width=\linewidth]{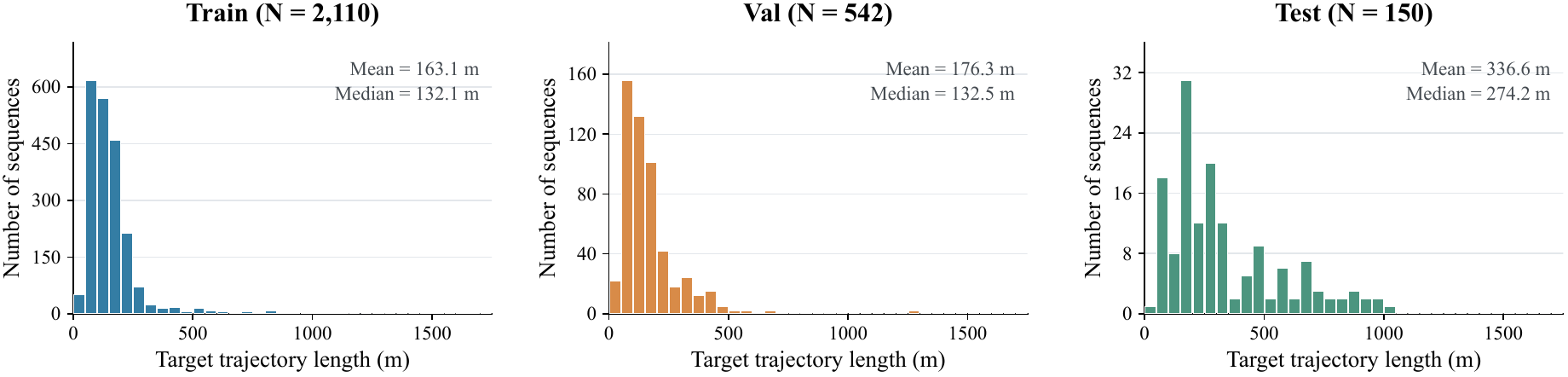}
    \caption{Target trajectory length distributions across the Train, Val, and Test splits. Each sequence contributes one path-length measurement, computed as the sum of Euclidean distances between consecutive recorded 3D target positions. All panels use the same 50-m bins and horizontal range; vertical axes show sequence counts with independently scaled limits. Mean and median lengths are annotated for each split.}
    \label{fig:trajectory_length}
\end{figure}

\subsection{Target trajectory length.}
Figure~\ref{fig:trajectory_length} summarizes the cumulative 3D travel distance of the designated target in each sequence,
\begin{equation}
    L = \sum_{t=0}^{N-2}\left\|\mathbf{p}_{t+1}-\mathbf{p}_t\right\|_2.
\end{equation}
Train and Val have similar median lengths of 132.1\,m and 132.5\,m, respectively, while Val has a larger upper tail (P95: 430.8\,m versus 341.6\,m for Train).
Test contains longer target trajectories, with a mean of 336.6\,m and a median of 274.2\,m, compared with mean lengths of 163.1\,m and 176.3\,m for Train and Val.
Thus, although all splits contain a range of travel distances, their trajectory-length distributions differ, with Test emphasizing longer target travel distances.

\begin{figure*}[t]
    \centering
    \includegraphics[width=\linewidth]{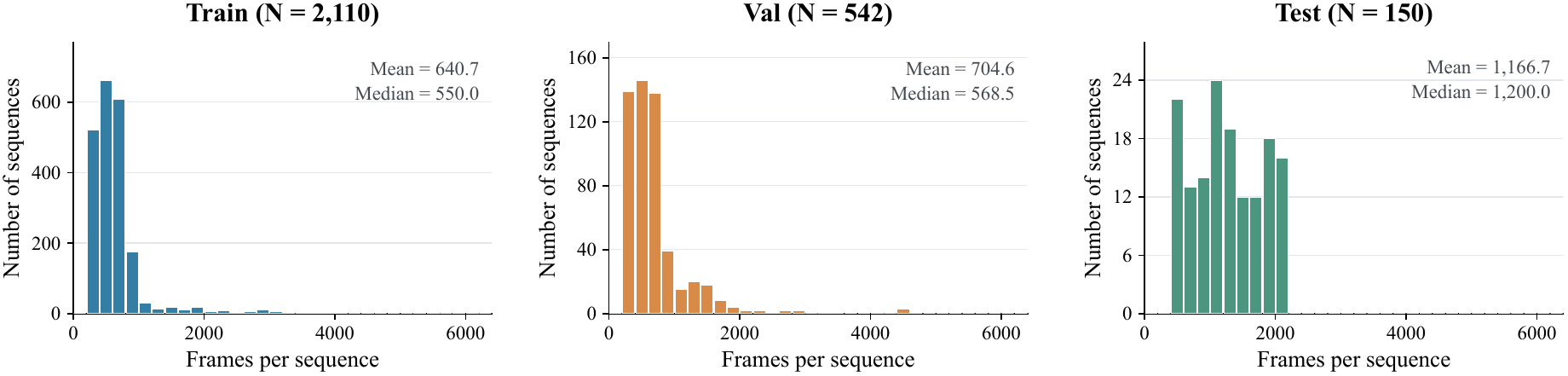}
    \caption{Distributions of recorded frame counts per sequence across the Train, Val, and Test splits. Each sequence contributes one observation. All panels use the same 200-frame bins and horizontal range, while vertical axes show sequence counts with independently scaled limits. Mean and median frame counts are annotated. Frames refer to synchronized ERP observations recorded at 10\,Hz.}
    \label{fig:sequence_frames}
\end{figure*}

\subsection{Sequence length in frames.}
Figure~\ref{fig:sequence_frames} summarizes the number of recorded ERP frames per sequence.
Train and Val have median lengths of 550 and 568.5 frames, respectively, with mean lengths of 640.7 and 704.6 frames.
Both splits contain longer sequences, reaching 6,000 frames in Train and 6,286 frames in Val.
Test has a higher median of 1,200 frames and a mean of 1,166.7 frames, with sequence lengths ranging from 400 to 2,000 frames.
Thus, typical Test sequences contain more observations, whereas Train and Val cover a wider range that extends to substantially longer recordings.

\begin{figure*}[t]
    \centering
    \includegraphics[width=0.9\linewidth]{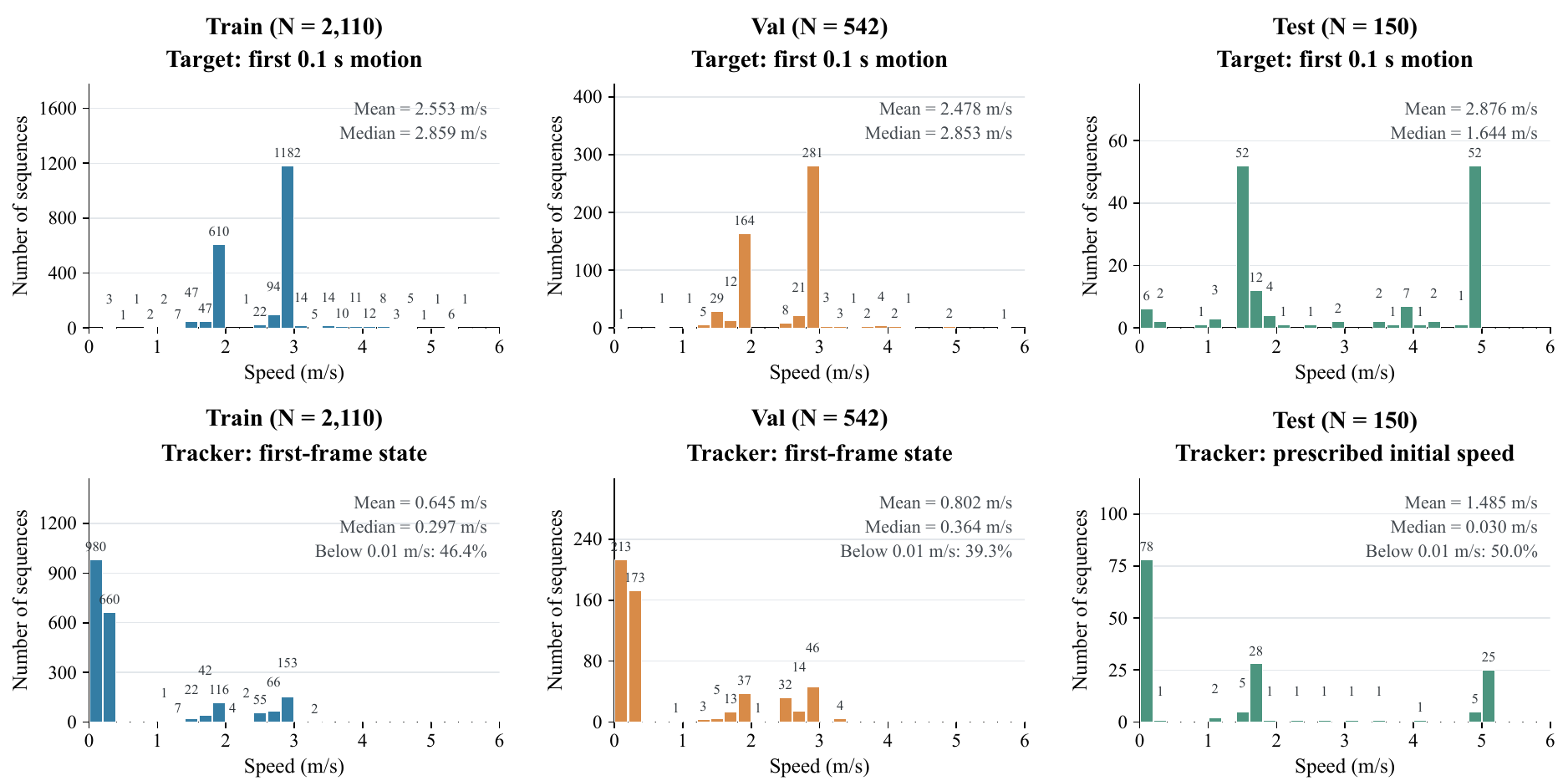}
    \caption{Initial motion and tracker initialization speed distributions across Train, Val, and Test. Top: target speed estimated from the first two historical expert 3D positions over a 0.1-s interval. Bottom: recorded first-frame tracker speed for Train and Val, and protocol-prescribed tracker initial speed for Test. Test contains 75 stationary and 75 moving initializations. Each sequence contributes one observation per actor. Numbers above nonempty bars indicate sequence counts in each bin. All panels use 0.2-m/s bins and a common horizontal range; vertical axes are independently scaled. Test initialization settings are distinguished from historical expert trajectory measurements.}
    \label{fig:initial_speed}
\end{figure*}

\paragraph{Initial motion and tracker states.}
Figure~\ref{fig:initial_speed} separates target motion at the start of a recorded expert sequence from tracker initialization speeds, using recorded first-frame tracker velocities for Train and Val and protocol-prescribed initial velocities for Test.
Since simulator-reported target velocities can be zero for actors whose recorded positions change, target initial motion is estimated as $s_{\mathrm{T},0}=\|\mathbf{p}_1-\mathbf{p}_0\|_2/(t_1-t_0)$.
This is the mean motion speed over the first recorded interval, not an instantaneous measurement at $t_0$.
Tracker initial speed is the norm of its recorded first-frame velocity for Train and Val, and the norm of the protocol-prescribed world-coordinate velocity vector for Test.
Mean target initial motion speeds are 2.553, 2.478, and 2.876\,m/s for Train, Val, and Test, respectively.
Mean tracker initialization speeds are 0.645, 0.802, 1.485\,m/s, respectively.
The fractions of tracker initial speeds below 0.01\,m/s are 46.4\%, 39.3\%, and 50.0\%, respectively.
The Test protocol explicitly balances stationary and moving initializations (75 sequences each).
For each moving initialization, the prescribed tracker velocity equals the target motion reference vector used by its controller; it need not equal the target's measured velocity at frame zero during settling or acceleration.
Tracker motion estimated from historical expert positions over the first interval is a separate quantity and does not represent the updated mixed online initialization.

\section{Additional Qualitative Analysis}
\label{sec:appendix_qualitative}


\begin{figure*}[t]
    \centering
    \includegraphics[width=0.9\linewidth]{
        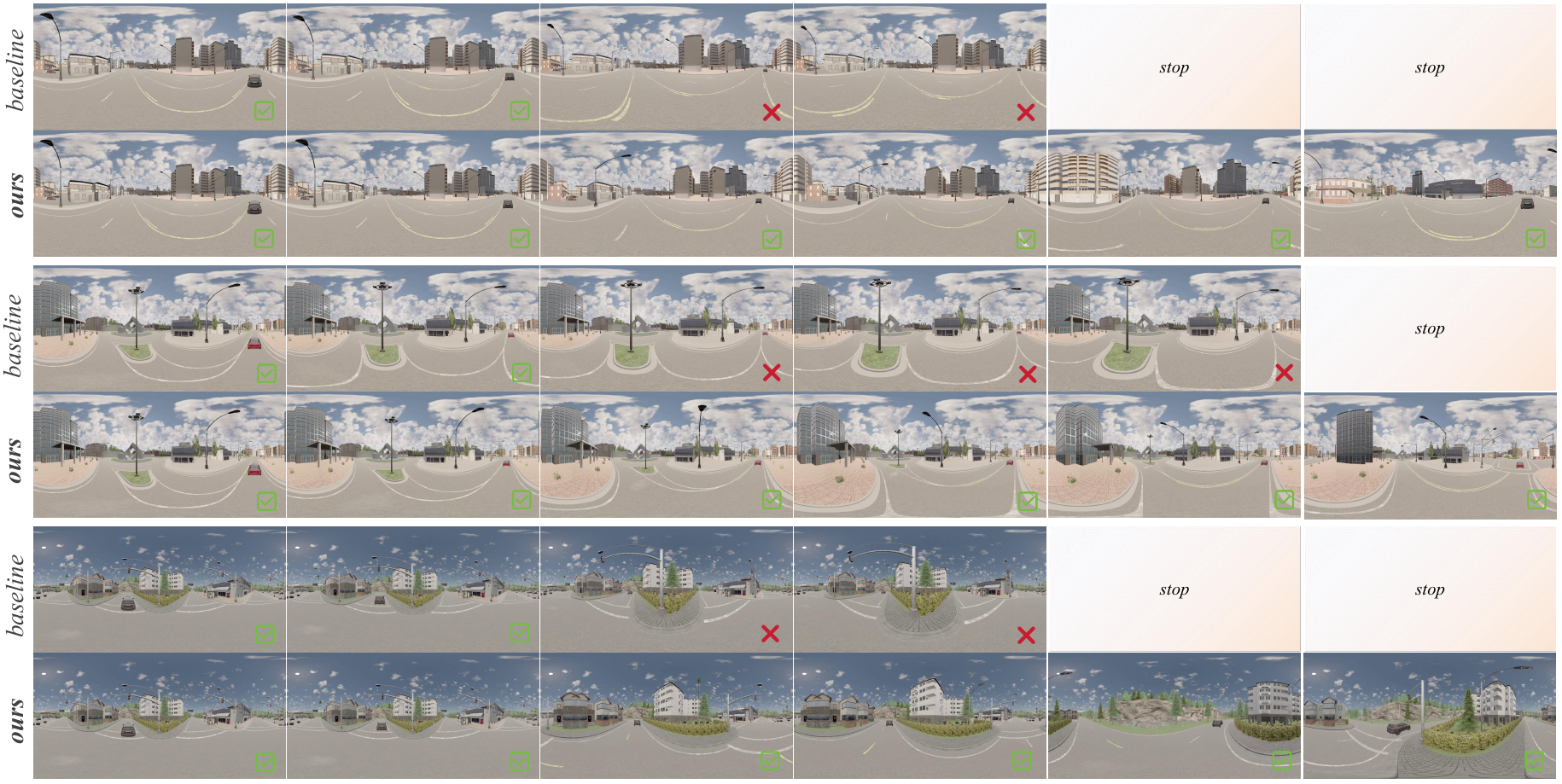}
    \caption{
    \textbf{Additional closed-loop tracking comparisons.}
    Three representative vehicle-following scenarios are
    shown over successive time steps.
    Each pair of rows compares the baseline and our
    360UAV-TVLA under full-spherical observation and
    fixed-heading control.
    Green checkmarks indicate valid tracking, red crosses
    indicate tracking failures, and ``stop'' denotes
    terminated rollouts.
    While the baseline progressively loses effective
    tracking and terminates early, our method maintains
    valid following throughout the illustrated sequences.
    }
    \label{fig:appendix_tracking}
\end{figure*}

\subsection{Additional closed-loop tracking comparisons.}
Figure~\ref{fig:appendix_tracking} presents three
additional closed-loop tracking sequences under
different urban driving conditions.
Both the baseline and our method initially maintain
valid tracking, but their behaviors gradually diverge
as the relative target configuration changes.
The baseline exhibits tracking failures and
premature termination, whereas 360UAV-TVLA
continues to maintain valid following.
Since both methods operate with full-spherical
observations and fixed-heading control, these
comparisons further illustrate the importance
of effectively translating panoramic perception
into continuous translational actions.
The evolving panoramic observations also demonstrate
the substantial viewpoint changes encountered
during closed-loop UAV motion.

\begin{figure}[t]
    \centering
    \includegraphics[width=1.0\linewidth]{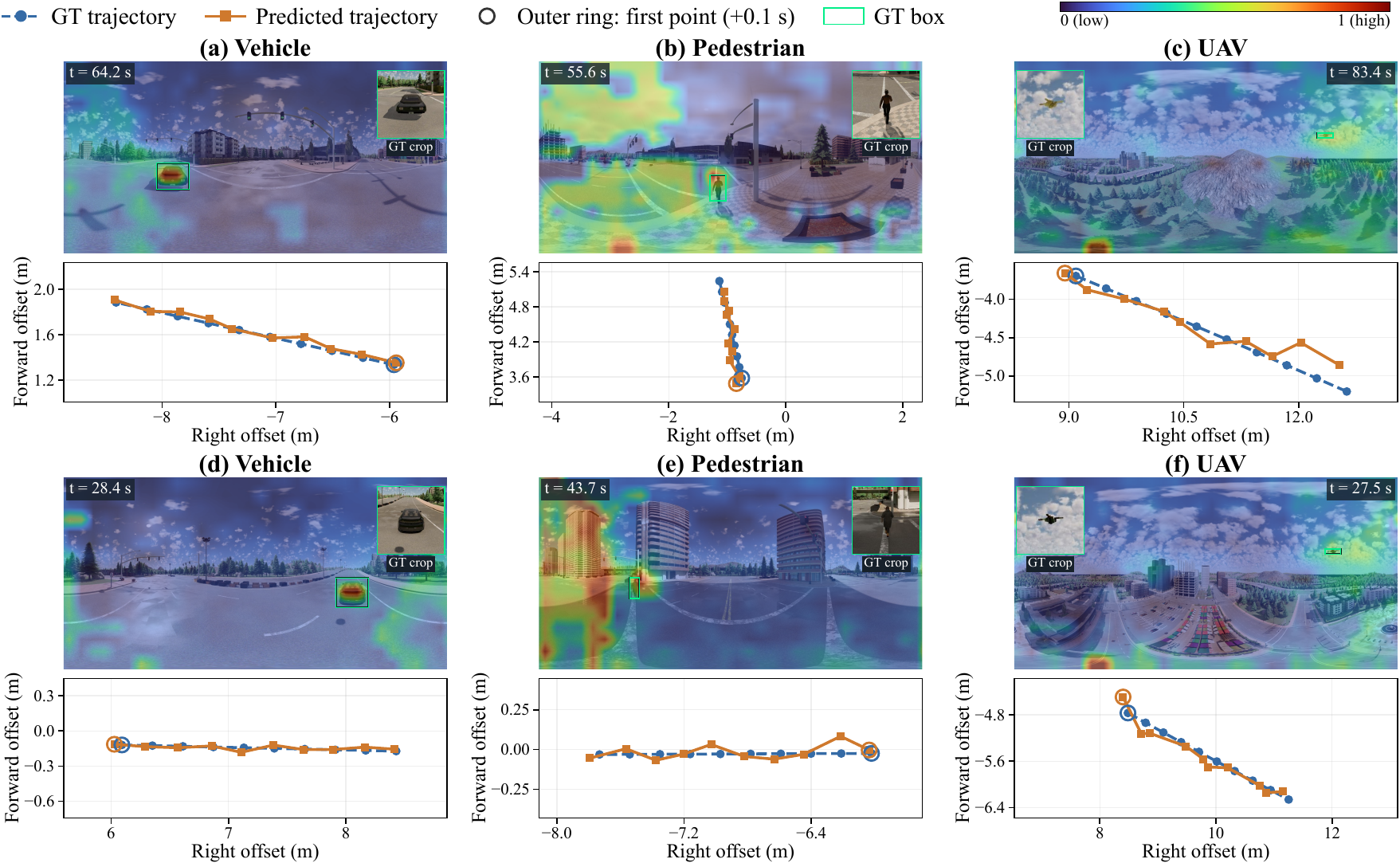}
    \caption{\textbf{Additional future-target predictions.}
    Horizontal projections over the next second for the first two
    complete-horizon Val samples in each target class. All points
    use the current UAV origin and initial-heading axes. Reported
    ADE/FDE use all three coordinates.}
    \label{fig:appendix_target_future}
\end{figure}

\begin{figure}[t]
    \centering
    \includegraphics[width=1.0\linewidth]{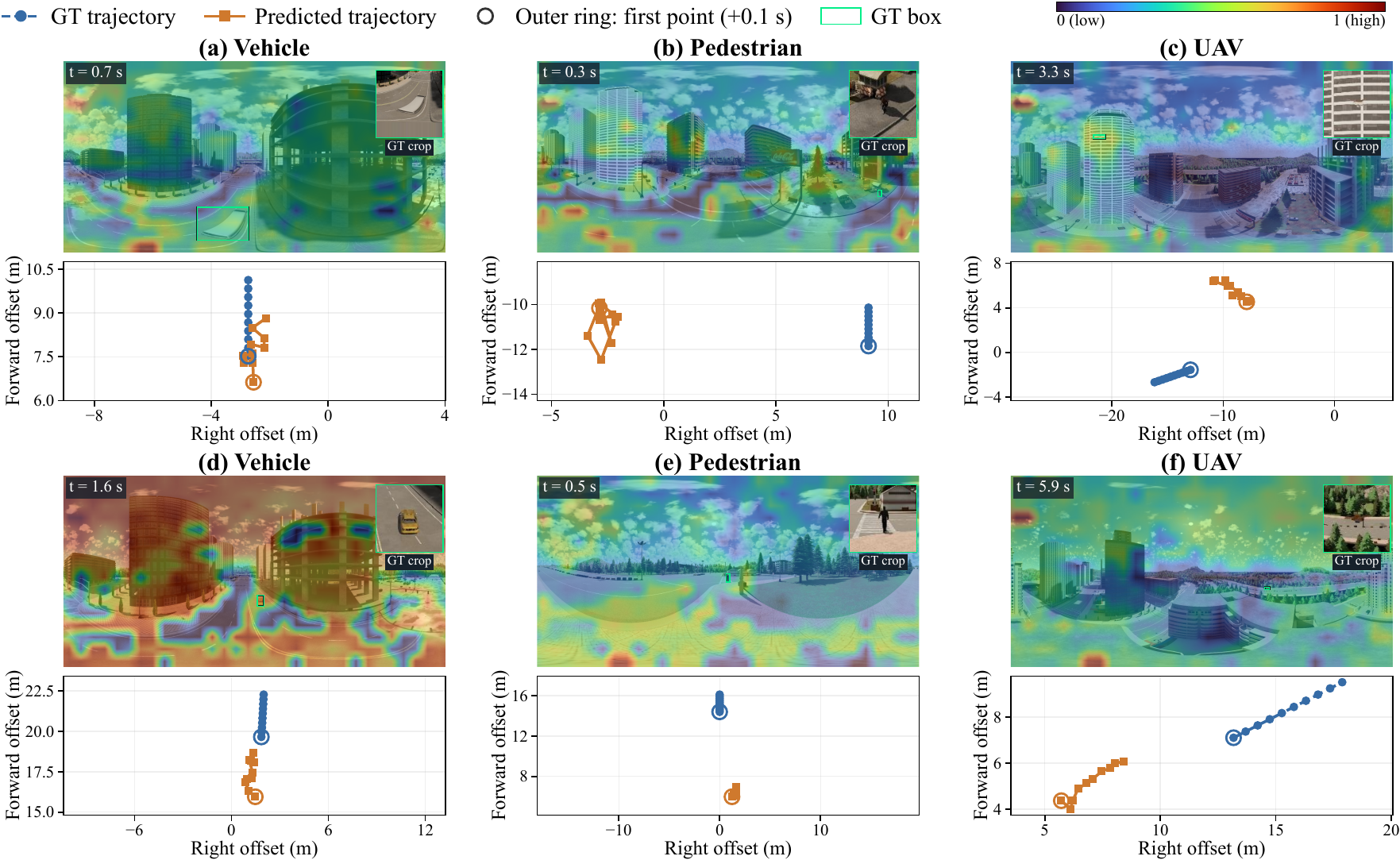}
    \caption{\textbf{Additional future-target predictions.}
    Horizontal projections over the next second for the first two
    complete-horizon Val samples in each target class. All points
    use the current UAV origin and initial-heading axes. Reported
    ADE/FDE use all three coordinates.}
    \label{fig:appendix_target_future_failed}
\end{figure}

\subsection{Future-target prediction and evaluation.}
Figure~\ref{fig:appendix_target_future} visualizes outputs of the
training-only trajectory head at the checkpoint selected by velocity
MAE. The examples follow the fixed sample order and are not selected
by forecast error. Dashed ground-truth and solid predicted paths
use the same current-UAV coordinate frame; connecting future points
does not depict a planned UAV path. The plots show horizontal
projections, while numerical errors include height.

For normalized head output $\widehat{\mathbf u}_{n,j}$, the physical
prediction is $\widehat{\mathbf q}_{n,j}=\boldsymbol\mu_q+
\boldsymbol\sigma_q\odot\widehat{\mathbf u}_{n,j}$.
Target ADE averages $\|\widehat{\mathbf q}_{n,j}-\mathbf q^\star_{n,j}\|_2$
over valid future points. Target FDE@1s uses $j=10$ only for samples
with the complete ten-step horizon. Occluded targets retain labels
when simulator positions are available. These metrics describe
short-horizon target prediction, distinct from the integrated UAV
trajectory ADE in Section~\ref{app:offline_ade}.
The head is bypassed during action-only inference and does not
provide coordinates to the controller.


\begin{figure*}[t]
    \centering
    \includegraphics[width=0.9\linewidth]{
        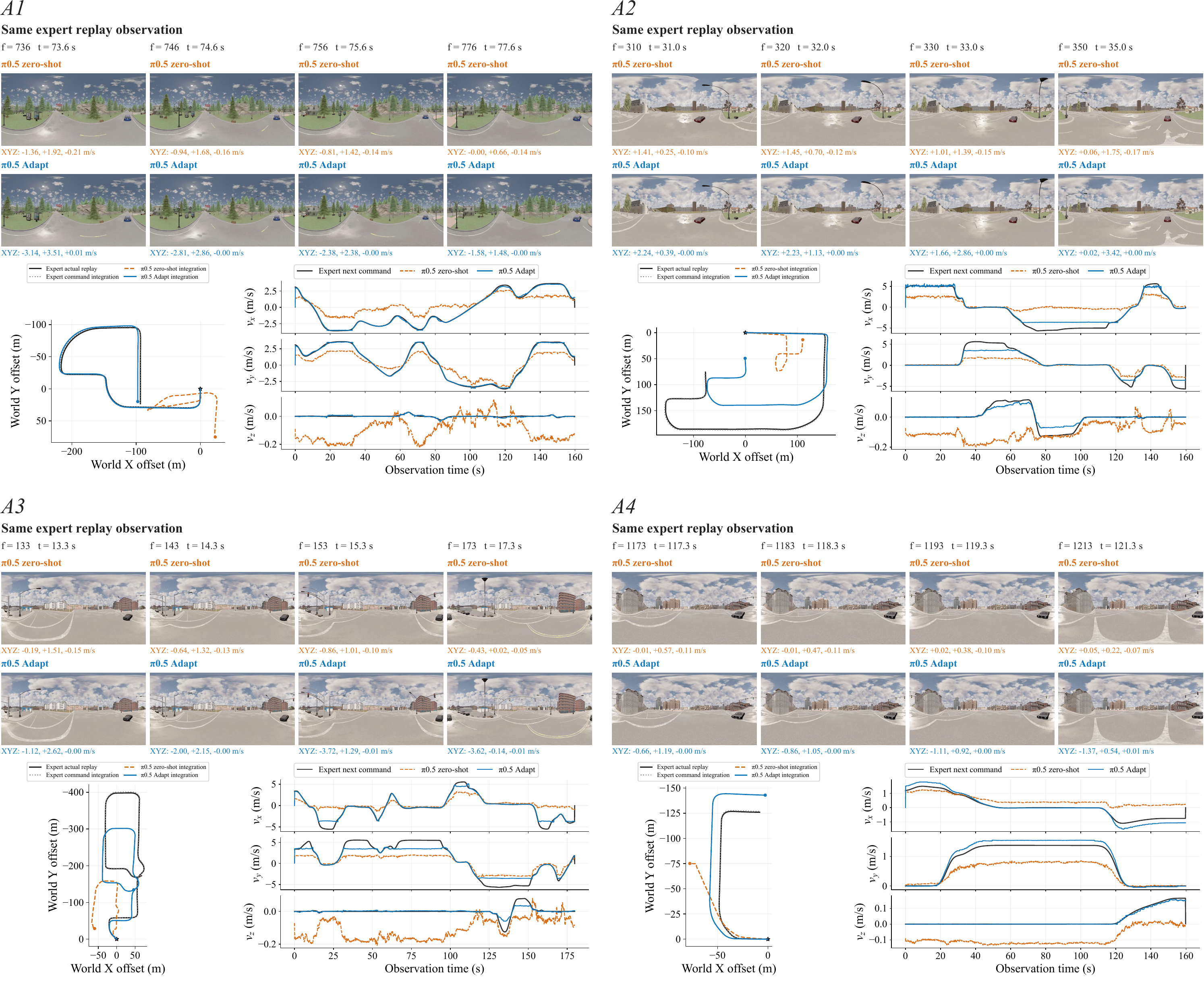}
    \caption{
    \textbf{Expert trajectory replay comparison on Task I.}
    Four front-initialized scenarios (A1--A4) compare
    zero-shot $\pi_{0.5}$ and its benchmark-adapted variant
    under identical recorded expert observations.
    The upper panels show sampled panoramic inputs
    and predicted XYZ velocity commands.
    The lower panels compare trajectories obtained
    by integrating the predicted commands with
    expert trajectories (left), together with
    the corresponding velocity components (right).
    Black, orange, and blue denote expert,
    zero-shot, and adapted predictions, respectively.
    All predicted trajectories are offline integrations,
    rather than actual closed-loop rollouts.
    }
    \label{fig:appendix_replay_task1}
\end{figure*}

\subsection{Expert trajectory replay on Task I.}
Figure~\ref{fig:appendix_replay_task1} compares
zero-shot $\pi_{0.5}$ with its adapted counterpart
across four front-initialized tracking scenarios.
Both policies receive the same recorded expert
observations, allowing their predicted actions
to be compared without changes in the observation
distribution induced by policy execution.
The zero-shot policy frequently produces velocity
commands that deviate from expert behavior,
leading to substantial differences in
integrated motion trajectories.
In comparison, the adapted policy generally
reproduces the direction and temporal variation
of expert commands more closely, particularly
during changes in translational velocity.
Nevertheless, noticeable trajectory deviations
remain in several scenarios, indicating that
action prediction errors may still accumulate
over extended sequences.
These results highlight the challenge of
transferring generalist action policies to
full-spherical UAV tracking.


\begin{figure*}[t]
    \centering
    \includegraphics[width=0.9\linewidth]{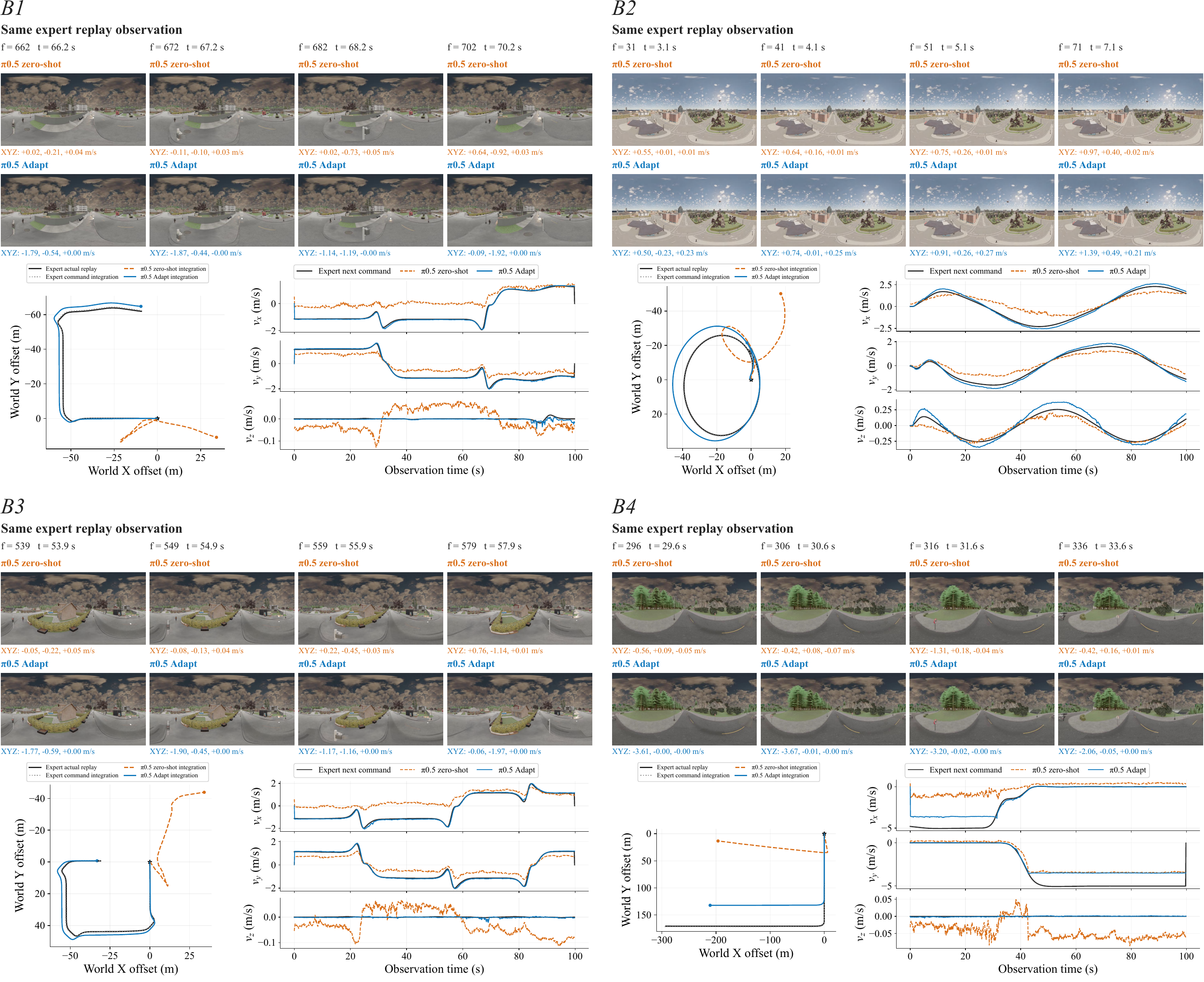}
    \caption{
    \textbf{Expert trajectory replay comparison on Task II.}
    Representative omnidirectional-initialization
    scenarios B1-B4 are evaluated using
    the same recorded expert observations.
    We compare zero-shot $\pi_{0.5}$ with its
    benchmark-adapted counterpart through
    predicted XYZ velocity commands and
    their integrated trajectories.
    The adapted policy exhibits closer agreement
    with expert motion in the illustrated examples,
    while residual prediction and trajectory errors
    remain in challenging cases.
    }
    \label{fig:appendix_replay_task2}
\end{figure*}

\subsection{Expert trajectory replay on Task II.}
Figure~\ref{fig:appendix_replay_task2} further
examines action prediction under omnidirectional
target initialization.
Unlike Task I, these scenarios require the policy
to reason about target directions that are not
restricted to the frontal viewing region.
The zero-shot policy often exhibits limited
responses to changes in expert translational
commands, resulting in inaccurate integrated
trajectories.
After adaptation, the predicted velocity
profiles more closely follow the expert commands
in the illustrated cases, especially during
changes in lateral and longitudinal motion.
However, the remaining discrepancies indicate
that accurate action prediction under arbitrary
target bearings remains challenging.
Together with Task I, these observations motivate
the need for geometry-aware perception and
temporally informed action prediction.

Importantly, the replay experiments in
Figures~\ref{fig:appendix_replay_task1}
and~\ref{fig:appendix_replay_task2}
measure action predictions along fixed expert
observation sequences.
They therefore provide diagnostics of offline
action imitation rather than direct evidence
of closed-loop tracking success.


\begin{figure}[t]
    \centering
    \includegraphics[width=1.0\linewidth]{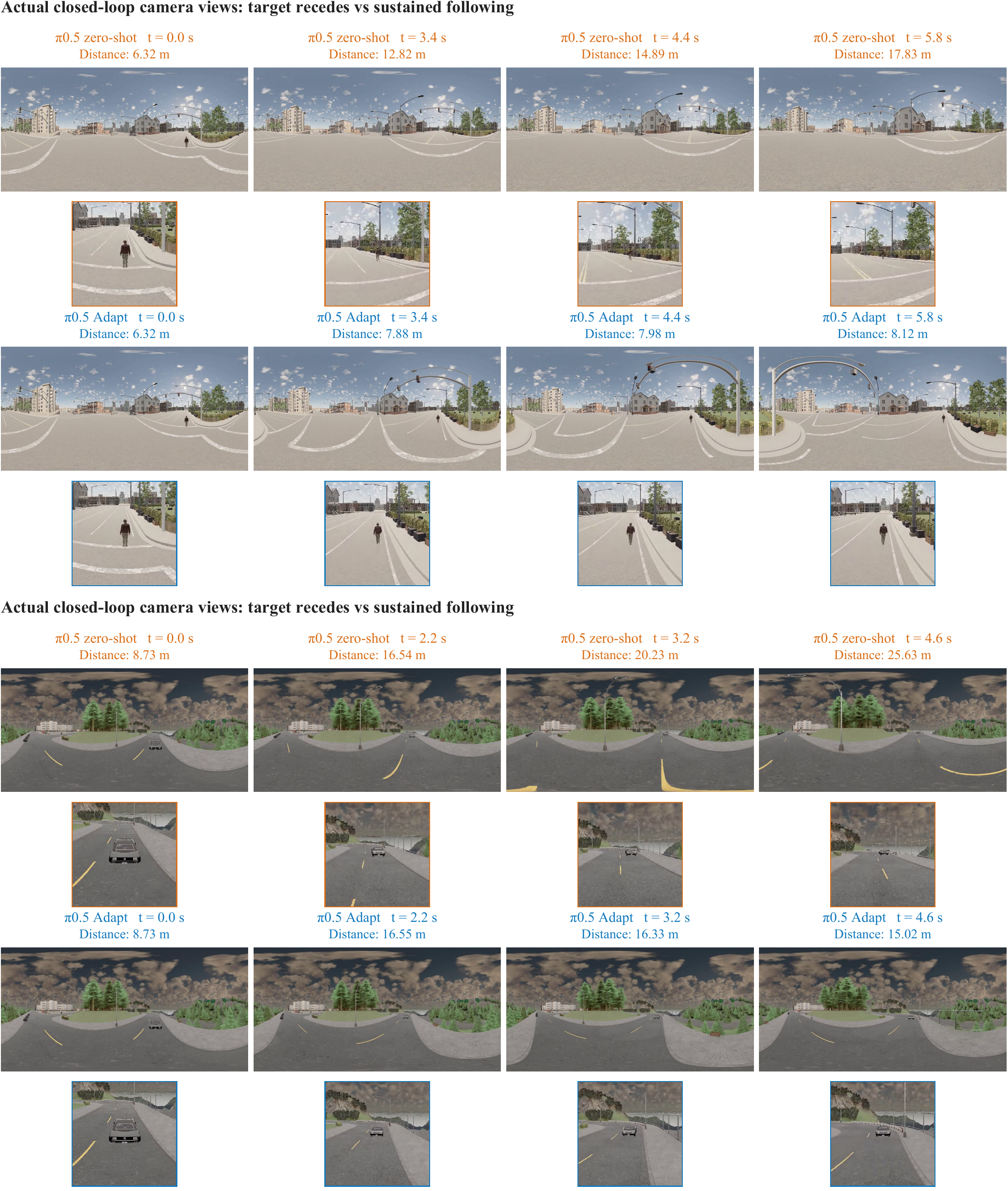}
    \caption{
    \textbf{Actual closed-loop comparison of
    zero-shot and adapted $\pi_{0.5}$.}
    Two representative episodes involving
    pedestrian following (top) and vehicle following
    (bottom) are shown at matched timestamps.
    For each policy, panoramic observations
    and corresponding perspective views
    illustrate the evolving target configuration.
    Target--UAV distances are annotated above
    the panoramic frames.
    Zero-shot $\pi_{0.5}$ exhibits increasing
    target separation, whereas the adapted policy
    maintains a smaller relative distance
    throughout the illustrated intervals.
    }
    \label{fig:appendix_pi05_closedloop}
\end{figure}

\subsection{Actual closed-loop behavior.}
To complement offline expert replay,
Figure~\ref{fig:appendix_pi05_closedloop}
compares the actual closed-loop behavior of
zero-shot and adapted $\pi_{0.5}$ policies.
Unlike the preceding replay experiments,
each policy executes its own predicted commands
and subsequently observes the resulting
simulation states.

In the pedestrian-following example,
the target distance under zero-shot $\pi_{0.5}$
increases from 6.32\,m to 17.83\,m
within 5.8\,s.
By contrast, the adapted policy maintains
a distance of 8.12\,m at the same timestamp.
A similar trend is observed in the vehicle-following
example, where the zero-shot distance increases
from 8.73\,m to 25.63\,m within 4.6\,s,
compared with 15.02\,m for the adapted policy.
These examples show that benchmark adaptation
can substantially improve short-term target
following under closed-loop execution.
They also demonstrate why expert-replay accuracy
alone is insufficient to characterize embodied
tracking performance, since policies must
continually react to the observations generated
by their own actions.

\section{Data Construction for the Observation--Control Ablation}
\label{app:abc_data_generation}

The three variants in
Table~\ref{tab:erp_fixed_heading_ablation} share the same expert
demonstrations, episode splits, target identities, instructions,
and timestamps.
The training and validation sets contain 2,110 and 542 episodes,
respectively, sampled at 10\,Hz.
Translational labels are identical across variants and expressed
in the original initial-heading coordinate frame, including
when C rotates.

\paragraph{Shared settings.}
All variants use the same static target-context reference from
the first ERP frame and three-frame history at
$[t-0.2,t-0.1,t]$, with missing history masked.
Each observation supervises the next ten actions starting at
$t+1$, with unavailable labels masked.
The training-only target-trajectory auxiliary loss has weight
2.0; target ground truth is never supplied as policy input.
For each backbone setting, all variants are independently initialized
from the same pretrained $\pi_0$ or $\pi_{0.5}$ checkpoint, using seed 42
and a fresh optimizer.
Training uses global batch size 64, 21,091 updates, learning rate
$2.5\times10^{-5}$, and XYZ normalization computed from the
shared training set.
Online evaluation uses identical episodes, original distance
bounds, a common ERP visibility criterion, and a
15-consecutive-invalid-frame failure threshold.
Leaving the perspective FoV alone does not terminate an episode.

\subsection{A: Limited FoV}
\label{app:abc_a}

\subsubsection{Data Construction}
Perspective images are geometrically reprojected from the same
ERP frames used by B, preserving timestamps and optical centers.
The virtual camera retains the original initial heading, with
a $90^\circ$ horizontal FoV, zero pitch, and
$960\times540$ resolution.
Projection uses bilinear interpolation and horizontal wrapping
at the ERP seam, followed by letterboxing to $448\times224$.
The original XYZ labels are retained.

\subsubsection{Training}
A predicts $(v_x,v_y,v_z)$ under the shared training settings.
It uses the same reference, history configuration, and
target-trajectory supervision as B, allowing the comparison
to focus on live observation coverage.
Checkpoint selection uses validation XYZ MAE.

\subsubsection{Closed-Loop Evaluation}
A executes its predicted XYZ velocities while maintaining the
original heading; the camera never automatically faces the target.
An unseen target can enter the perspective FoV only through
relative motion.
All scoring and termination rules remain identical to B.

\subsection{B: Fixed-Heading ERP}
\label{app:abc_b}

\subsubsection{Data Construction}
B uses the original ERP frames resized to $448\times224$,
with the same translational labels as A and C.

\subsubsection{Training}
B follows the shared training settings and selects its checkpoint
by validation XYZ MAE.

\subsubsection{Closed-Loop Evaluation}
B predicts $(v_x,v_y,v_z)$ with a fixed heading and receives
$360^\circ$ ERP observations.
It serves as the panoramic counterpart to A under the same
demonstration budget and evaluation protocol.

\begin{table*}[t]
\centering
\caption{
\textbf{Online closed-loop tracking under Protocol (i).}
The simulator is paused during the complete observation,
transport, and model-inference pipeline. ATF is reported in frames,
normalized ATF and SR in \%.
}
\vspace{-1em}
\label{tab:protocol_i_tracking}
\small
\setlength{\tabcolsep}{4pt}
\renewcommand{\arraystretch}{1.00}

\resizebox{\linewidth}{!}{%
\begin{tabular}{@{}lcc*{9}{c}@{}}
\toprule
\multirow{2}{*}{Method}
& \multirow{2}{*}{Publication}
& \multirow{2}{*}{Type}
& \multicolumn{3}{c}{\shortstack[c]{Task I Front-Initialized}}
& \multicolumn{3}{c}{\shortstack[c]{Task II Omnidirectional}}
& \multicolumn{3}{c}{\shortstack[c]{Overall Both Tasks}} \\
\cmidrule(lr){4-6}
\cmidrule(lr){7-9}
\cmidrule(lr){10-12}
& & &
ATF$\uparrow$
& $\mathrm{ATF}_{\mathrm{norm}}\uparrow$
& $\mathrm{SR}_{\mathrm{on}}\uparrow$
& ATF$\uparrow$
& $\mathrm{ATF}_{\mathrm{norm}}\uparrow$
& $\mathrm{SR}_{\mathrm{on}}\uparrow$
& ATF$\uparrow$
& $\mathrm{ATF}_{\mathrm{norm}}\uparrow$
& $\mathrm{SR}_{\mathrm{on}}\uparrow$ \\

\midrule

Temporal BC
& Baseline & BC
& 149.10 & 12.62 & 0.00
& 98.67 & 11.80 & 2.02
& 115.62 & 12.12 & 0.67 \\

D-VAT-BC~\citep{dionigi2024dvat}
& RA-L'24 & VAT
& 171.33 & 14.00 & 1.96
& 85.66 & 10.90 & 1.01
& 114.66 & 11.92 & 1.33 \\

Fast-WAM~\citep{yuan2026fastwam}
& NeurIPS'26 & WAM
& 170.16 & 14.54 & 1.96
& 178.66 & 20.69 & 3.03
& 175.77 & 18.60 & 2.67 \\

$\pi_0$ (w/ ft)~\citep{black2025pi0}
& RSS'25 & VLA
& 310.73 & 23.90 & 3.92
& 188.30 & 21.87 & 6.06
& 224.24 & 22.48 & 6.00 \\

$\pi_{0.5}$ (w/ ft)~\citep{black2025pi05}
& CoRL'25 & VLA
& 305.55 & 24.40 & 5.88
& 212.61 & 26.34 & 7.07
& 247.92 & 26.30 & 6.00 \\

\midrule

\rowcolor{blue!4}
\textbf{360UAV-TVLA ($\pi_0$)}
& -- & \textbf{VLA}
& \textbf{568.10} & \textbf{40.74} & \textbf{9.80}
& \textbf{304.54} & \textbf{34.26} & \textbf{13.13}
& \textbf{394.15} & \textbf{36.46} & \textbf{12.00} \\

\rowcolor{blue!4}
\textbf{360UAV-TVLA ($\pi_{0.5}$)}
& -- & \textbf{VLA}
& \underline{429.69} & \underline{32.68} & \textbf{9.80}
& \underline{290.69} & \underline{33.95} & \textbf{13.13}
& \underline{337.95} & \underline{33.52} & \textbf{12.00} \\

\bottomrule
\end{tabular}%
}
\vspace{-1.2em}
\end{table*}
\begin{table*}[t]
\centering
\caption{
\textbf{Online closed-loop tracking under Protocol (ii).}
The simulator continues running during model inference, while the UAV
holds its previous command until the returned action is applied at the
next physical boundary. ATF is reported in frames, normalized ATF and
SR in \%.
}
\vspace{-1em}
\label{tab:protocol_ii_tracking}
\small
\setlength{\tabcolsep}{4pt}
\renewcommand{\arraystretch}{1.00}

\resizebox{\linewidth}{!}{%
\begin{tabular}{@{}lcc*{9}{c}@{}}
\toprule
\multirow{2}{*}{Method}
& \multirow{2}{*}{Publication}
& \multirow{2}{*}{Type}
& \multicolumn{3}{c}{\shortstack[c]{Task I Front-Initialized}}
& \multicolumn{3}{c}{\shortstack[c]{Task II Omnidirectional}}
& \multicolumn{3}{c}{\shortstack[c]{Overall Both Tasks}} \\
\cmidrule(lr){4-6}
\cmidrule(lr){7-9}
\cmidrule(lr){10-12}
& & &
ATF$\uparrow$
& $\mathrm{ATF}_{\mathrm{norm}}\uparrow$
& $\mathrm{SR}_{\mathrm{on}}\uparrow$
& ATF$\uparrow$
& $\mathrm{ATF}_{\mathrm{norm}}\uparrow$
& $\mathrm{SR}_{\mathrm{on}}\uparrow$
& ATF$\uparrow$
& $\mathrm{ATF}_{\mathrm{norm}}\uparrow$
& $\mathrm{SR}_{\mathrm{on}}\uparrow$ \\

\midrule

Temporal BC
& Baseline & BC
& 184.96 & 15.28 & 1.96
& \textbf{218.33} & \textbf{24.79} & \textbf{6.06}
& \textbf{206.99} & \textbf{21.56} & \textbf{4.67} \\

D-VAT-BC~\citep{dionigi2024dvat}
& RA-L'24 & VAT
& 188.35 & 14.96 & 1.96
& 174.21 & 19.90 & 2.02
& 179.02 & 18.22 & 2.00 \\

Fast-WAM~\citep{yuan2026fastwam}
& NeurIPS'26 & WAM
& 177.16 & 14.90 & 1.96
& 182.21 & 21.17 & 3.03
& 180.49 & 19.04 & 2.67 \\

$\pi_0$ (w/ ft)~\citep{black2025pi0}
& RSS'25 & VLA
& \textbf{195.31} & \textbf{16.10} & 1.96
& 169.04 & 19.57 & 3.03
& 177.97 & 18.39 & 2.67 \\

$\pi_{0.5}$ (w/ ft)~\citep{black2025pi05}
& CoRL'25 & VLA
& 177.39 & 15.41 & \textbf{3.92}
& 177.11 & 20.50 & 2.02
& 177.21 & 19.43 & 2.67 \\

\midrule

\rowcolor{blue!4}
\textbf{360UAV-TVLA ($\pi_0$)}
& -- & \textbf{VLA}
& 160.27 & 13.84 & 1.96
& \textbf{220.45} & 24.81 & \underline{4.04}
& \underline{199.99} & \underline{21.08} & \underline{3.33} \\

\rowcolor{blue!4}
\textbf{360UAV-TVLA ($\pi_{0.5}$)}
& -- & \textbf{VLA}
& 152.88 & 13.40 & 1.96
& 201.60 & \underline{22.06} & \underline{4.04}
& 185.03 & 19.12 & \underline{3.33} \\

\bottomrule
\end{tabular}%
}
\vspace{-1.2em}
\end{table*}

\subsection{C: Active Steering}
\label{app:abc_c}

\subsubsection{Data Construction}
C uses the same perspective projection as A, but with a
time-varying heading.
XYZ labels remain unchanged; yaw-rate labels are generated
by a causal, visibility-constrained expert along the recorded
translational trajectory.
The expert accesses the target bearing only when the target
is annotated as visible in ERP and lies within the current
perspective frustum.
For a visible target,
\begin{equation}
    \omega_t^{\mathrm{des}}
    = 4\,\mathrm{s}^{-1}
      \operatorname{wrap}(\beta_t-\psi_t),
\end{equation}
where $\beta_t$ and $\psi_t$ denote target bearing and camera
heading, respectively, and angular differences are wrapped
to $[-\pi,\pi)$.
When the target is unseen, the expert turns toward its last
observed bearing, then sweeps at $120^\circ/\mathrm{s}$ once
within $5^\circ$.
Without a previous observation, it starts a positive-yaw sweep.
Neither current hidden target bearings nor future target
positions are used to select yaw commands.

Yaw rate and acceleration are limited to
$180^\circ/\mathrm{s}$ and $720^\circ/\mathrm{s}^{2}$.
Starting from the original heading and zero yaw rate,
the heading is updated as
\begin{equation}
    \psi_{t+1}=\psi_t+\Delta t\,\omega_{t+1},
    \qquad \Delta t=0.1\,\mathrm{s}.
\end{equation}
Each historical frame is projected using its own heading.
This construction adds steering supervision without collecting
additional demonstrations or changing the translational labels.

\subsubsection{Training}
C predicts $(v_x,v_y,v_z,\omega_{\mathrm{yaw}})$ and additionally
receives relative-heading sine/cosine and measured yaw rate.
It receives no target positions, visibility labels, or expert
search modes.
The backbone, training budget, reference, history configuration,
and target-trajectory supervision match A and B.
The action loss additionally includes normalized yaw error:
\begin{equation}
    \mathcal{L}_{\mathrm{act}}^{C}
    =
    \operatorname{MAE}_{XYZ}
    +
    \frac{\operatorname{MAE}_{\omega}}
         {\pi\,\mathrm{rad/s}}.
\end{equation}
Checkpoint selection uses the corresponding combined validation
score.
Optional yaw-recovery augmentation is excluded to preserve
the shared sample budget.

\subsubsection{Closed-Loop Evaluation}
C executes both predicted translation and yaw commands.
Perspective observations are generated from the actual simulated
heading, while XYZ commands remain in the original
initial-heading frame.
Episodes and scoring rules match A and B.
The training views synthesize camera rotations along recorded
translational trajectories, rather than using additional joint
translation--yaw demonstrations; thus, the comparison holds
the source demonstration budget fixed.

\subsubsection{Fairness of Active-Steering Supervision}
To ensure a fair comparison, we explicitly account for
non-frontal target initialization in both training and
closed-loop evaluation. When the target is initially outside
the perspective FoV, the steering expert performs a causal
yaw search until the target becomes visible, after which
it adjusts the heading to maintain visual contact.
This behavior is incorporated into the 4-DoF training
data generated from the original expert trajectories.
Although we do not recollect jointly optimized 4-DoF
expert demonstrations, the generated supervision provides
meaningful target-search and following behaviors while
preserving the original translational trajectories and
demonstration budget. During online evaluation, the
policy independently predicts and executes yaw commands
without access to privileged target information.

\subsection{Timing-protocol sensitivity.}
\label{app:timing_protocol_analysis}

Protocol (i) evaluates synchronized action application by pausing the
simulator during inference. Under this setting, 360UAV-TVLA achieves
the strongest overall performance among the evaluated VLA models.
The $\pi_0$ variant reaches 394.15 frames, 36.46\% normalized ATF,
and 12.00\% SR, while the $\pi_{0.5}$ variant reaches 337.95 frames,
33.52\% normalized ATF, and 12.00\% SR. Both variants outperform
their corresponding fine-tuned baselines in ATF, normalized ATF, and
success rate.
Protocol (ii) removes the synchronization barrier: the simulator
continues to evolve during inference and the UAV holds its previous
command until a new action is returned. Overall performance decreases
to 199.99 frames, 21.08\% normalized ATF, and 3.33\% SR for the
$\pi_0$ variant, and to 185.03 frames, 19.12\% normalized ATF, and
3.33\% SR for the $\pi_{0.5}$ variant. This reduction is attributed
to observation--action staleness caused by continued target and UAV
motion during inference. Nevertheless, both variants remain above
their corresponding fine-tuned $\pi$ baselines in overall ATF and SR.

The two protocols therefore measure complementary properties.
Protocol (i) isolates policy quality under synchronized action
application, whereas Protocol (ii) evaluates robustness to temporal
misalignment and delayed action updates. Protocol (ii) follows the
continued-world definition specified for this study; its clock
diagnostics are retained separately and are not used as an additional
score-eligibility criterion.


\end{document}